\documentclass[runningheads]{llncs}

\usepackage{eccv}

\usepackage{eccvabbrv}

\usepackage{graphicx}
\usepackage{booktabs}
\usepackage{multirow}

\usepackage{arydshln}
\usepackage[accsupp]{axessibility}

\usepackage[hidelinks]{hyperref}
\usepackage{orcidlink}

\newcommand{\rev}[1]{#1}

\begin{document}

\title{Video Generation Models Are Inherent \\ Lighting Estimators}

\titlerunning{Video Generation Models Are Inherent Lighting Estimators}

\author{Ziqi~Cai\inst{1,3} \and
Shuchen~Weng\inst{2,1}\textsuperscript{*} \and
Kaiqi~Liu\inst{1} \and
Zifeng~Wang\inst{1} \and\texorpdfstring{\\[-0.1em]}{ }
Zhiquan~Zhang\inst{1} \and
Minggui~Teng\inst{1} \and
Han~Jiang\inst{3} \and
Boxin~Shi\inst{1}\textsuperscript{*}}

\authorrunning{Z.~Cai et al.}
\tocauthor{Ziqi Cai, Shuchen Weng, Kaiqi Liu, Zifeng Wang, Zhiquan Zhang, Minggui Teng, Han Jiang, and Boxin Shi}

\institute{$^1$\,Peking University \quad
$^2$\,Beijing Academy of Artificial Intelligence\\
$^3$\,OpenBayes Information Technology Co., Ltd.\\
\email{czq@stu.pku.edu.cn, \{shuchenweng, shiboxin\}@pku.edu.cn}}

\maketitle
\begingroup
\renewcommand{\thefootnote}{}
\footnotetext[1]{$^{1}$~Ziqi Cai, Shuchen Weng, Zifeng Wang, Zhiquan Zhang, Minggui Teng, and Boxin Shi are with State Key Laboratory of Multimedia Information Processing and National Engineering Research Center of Visual Technology, School of Computer Science, Peking University. Kaiqi Liu is with School of Software and Microelectronics, Peking University.
\par\noindent $^{*}$ Corresponding authors.}
\endgroup

\begin{abstract}
Recovering dynamic environment maps from a single in-the-wild video is crucial for photorealistic rendering, yet remains a challenge. Recent video generation models can produce photorealistic scenes with complex lighting, possessing an inherent understanding of lighting. In this paper, we introduce V-LITE (Video generation models are inherent lighting estimators), a framework that unlocks this internal knowledge by reframing lighting estimation as a guided video inpainting task. Inspired by visual effects (VFX) industry practices, we insert a synthetic chrome ball into the scene to compel the model to generate physically plausible reflections from the surrounding spatio-temporal context. To bridge the gap from LDR-native models to the HDR domain, we design an HDR-aware VAE and employ an efficient LoRA-based fine-tuning strategy.
We then construct a mixed dataset comprising high-fidelity HDR images to provide realistic HDR priors, and in-the-wild HDR videos to provide dynamic spatio-temporal context.
Extensive experiments demonstrate that V-LITE produces temporally coherent HDR environment maps, revealing that modern video diffusion models are not merely synthesizers but also powerful, inherently capable estimators of physical scene lighting.


\end{abstract}

\section{Introduction}

As the primary factor to determine the visual appearance of in-the-wild scenes, an accurate lighting representation is the foundational cornerstone for photorealistic augmented reality~\cite{debevec2008rendering}, virtual object insertion~\cite{liang2024dipir}, and scene relighting~\cite{rao2024lite2relight,cai2024portrait}. However, recovering the complete environment map from a single in-the-wild video remains a significant challenge, due to the inherently partial illumination observations and the complex interactions among camera motion, object occlusions, and diverse material properties.

Classical physics-based methods~\cite{li2023hdrvideo} typically solve a simplified problem, assuming videos have static lighting, Lambertian surfaces, or known geometry.
Meanwhile, multi-view approaches~\cite{srinivasan2020lighthouse} solve the problem by introducing external guidance (\eg, images from additional perspectives). However, both paradigms fundamentally deviate from the intended task of processing an in-the-wild video. In contrast, recent work \cite{liang2025luxdit} feeds videos as conditions to generate low dynamic range (LDR) results for high dynamic range (HDR) fusion. While this approach presents a practical solution, it treats the model as a simple translator, ignoring its \textit{inherent} generative priors for illumination and thus potentially producing implausible results, such as exposure-fusion artifacts, color bias, and temporally unstable reflections.

\begin{figure}[t]
    \centering
    \includegraphics[width=\textwidth]{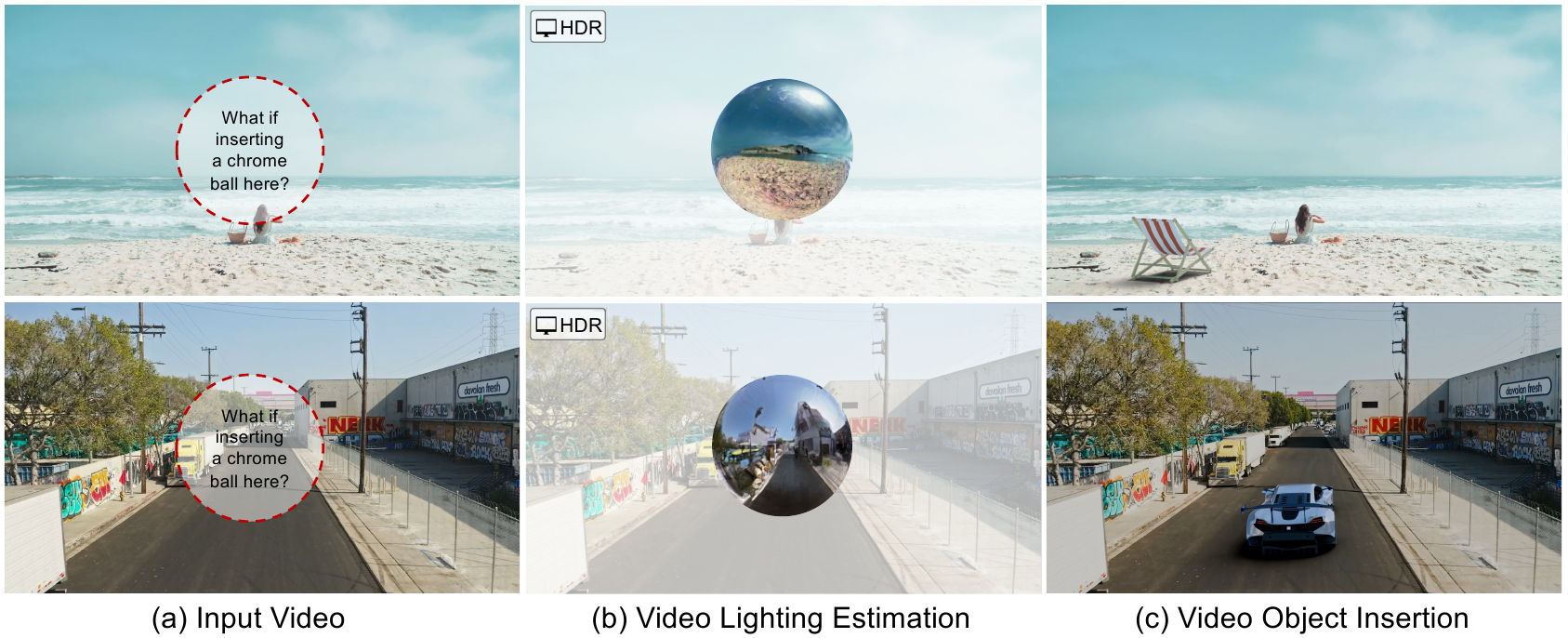}
    \captionsetup{type=figure}
    \captionof{figure}{{
    An illustration of our V-LITE framework.
    (\textbf{a}) Inspired by on-set VFX practices, we reformulate dynamic lighting estimation as the task of inserting a virtual light probe into any in-the-wild LDR video.
    (\textbf{b}) We implement this concept as a video inpainting task, compelling the video diffusion model to leverage its inherent priors and render a physically plausible chrome ball that captures the dynamic illumination of the scene.
    (\textbf{c}) The resulting HDR environment map enables downstream applications, such as inserting virtual objects with realistic and temporally consistent lighting and shadows.
    }}
    \label{fig:teaser}
\end{figure}

Recent video generation models~\cite{yang2024cogvideox,hong2022cogvideo, gao2025seedance, wan2025wan, cai2026lighting} have demonstrated an impressive capability to produce photorealistic scenes with complex light and shadow interactions, which highlights their implicit understanding of illumination and suggests an inherent capability for lighting estimation. This motivates us to explore these capabilities to directly generate HDR environment maps. However, extending existing video generation frameworks for lighting estimation introduces two key challenges: \textit{(i)} adapting the model to handle HDR processing, and \textit{(ii)} enabling HDR representations within the latent diffusion model.

In this paper, we introduce V-LITE (\textbf{V}ideo generation models are inherent \textbf{LI}gh\textbf{T}ing \textbf{E}stimators), a unified HDR-integrated framework that estimates environment maps directly from a single in-the-wild video.
We are inspired by the foundational practice in the visual effects industry, where a light probe (typically a chrome ball) is inserted into a scene to capture the complete surrounding illumination~\cite{murmann2019dataset,phongthawee2024diffusionlight}. This enables rendering virtual objects with photorealistic lighting consistency. Therefore, we reformulate the lighting estimation problem as a video inpainting task, requiring the model to fill in a synthetic chrome ball, which is then unwrapped to produce the final environment map. This approach forces the video generation model to consider both spatial context and temporal consistency when plausibly inserting the ball, thereby \textit{inherently} leveraging its generative priors. As shown in \cref{fig:teaser}, our method robustly generates high-quality environment maps even in extreme cases (\eg, fast camera motion).

We build our framework upon a diffusion-based network for LDR video generation~\cite{wan2025wan}. To process HDR videos with original LDR Variational AutoEncoder (VAE), we design and integrate learnable tonemap adapters, ensuring the fine-grained illumination cues (\eg, color temperature variations across time) are faithfully preserved in the latent space.
To further align the pretrained latent space with these HDR latents, we introduce a LoRA-based fine-tuning scheme~\cite{hu2021lora}. This effectively enhances the lighting understanding while preserving its original generative priors for spatial and temporal context.
To facilitate the model training and evaluation, we also construct V-LITESet, a dataset including over 8K HDR video pairs with dynamic temporal lighting and 800 static HDR images, totaling over 648K frames for effective co-training and evaluation.

Our contributions can be summarized as follows:
\begin{itemize}
\item We reformulate video lighting estimation as a light probe inpainting task, producing a dynamic environment map from a single in-the-wild video.

\item We design an HDR-aware VAE to preserve lighting cues and a LoRA-based fine-tuning scheme to align the latent space with HDR representations.

\item We construct a hybrid dataset comprising 8K videos with dynamic temporal lighting and 800 images with diverse luminance distributions, facilitating robust model training and comprehensive evaluation.

\end{itemize}

\section{Related Work}

\subsection{Lighting Estimation from Images}
Estimating scene illumination from a single image has been studied extensively in both vision and graphics. Early deep methods directly regress HDR indoor environment maps from limited-FoV LDR photos, showing that plausible light probes can be inferred without explicit geometry or material supervision~\cite{2017LavalIndoor}. Subsequent works decompose the task into geometry/visibility completion and LDR-to-HDR mapping to better preserve high-frequency cues and material-dependent effects~\cite{song2019neuralillumination, zhan2021emlight}. To improve spatial fidelity, fast indoor lighting estimation is explored~\cite{garon2019fastsv}, while HDR panorama generation enables lighting editing and robust HDR mapping~\cite{wang2022stylelight,tang2023luminaire,cai2025physedit}. Neural inverse rendering provides complementary, physics-grounded priors by jointly reasoning about geometry, BRDFs, and illumination from a single image~\cite{sengupta2019nir}. For outdoor scenes, single-image methods often predict parametric sun/sky lighting~\cite{holdgeoffroy2017deepoutdoor}.
Despite these advances, single-image approaches typically lack \emph{temporal coupling} and rarely provide \emph{native HDR} training/inference under unconstrained conditions.

\subsection{Lighting Estimation from Videos}

Moving from images to videos introduces the challenge of maintaining temporal consistency while still modeling dynamic illumination. Early progress leverages multi-view or stereo cues to infer lighting volumes with 3D coherence~\cite{srinivasan2020lighthouse}. Closer to our setting, Li~\etal~\cite{li2023hdrvideo} represent video illumination as a spatiotemporally consistent spherical-Gaussian lighting volume and train recurrent priors to suppress flicker and enforce cross-frame consistency for indoor HDR lighting.
Recent work further studies spatiotemporally consistent indoor lighting estimation directly from in-the-wild videos, emphasizing smoothness of predicted lighting fields under changing viewpoints and local light variations~\cite{tong2025videolighting}.
In parallel, sequence-based HDR lighting reconstruction methods encode spatially-varying illumination as Gaussian splats from image sequences~\cite{bolduc2025gaslight},
but typically require controlled capture settings and do not target generic casually captured videos.
Methods that aim at lightweight 3D-coherent representations can also improve spatial consistency (\eg, LightOctree), but they are primarily designed for single-image inputs and do not directly couple predictions across time~\cite{wang2024lightoctree}.
However, robust estimation for \emph{in-the-wild} videos that is simultaneously \emph{temporally consistent} and supports \emph{native HDR} remains under-explored, which we address by leveraging video diffusion priors and task-specific adaptation.

\begin{figure}[t]
    \centering
    \includegraphics[width=1\linewidth]{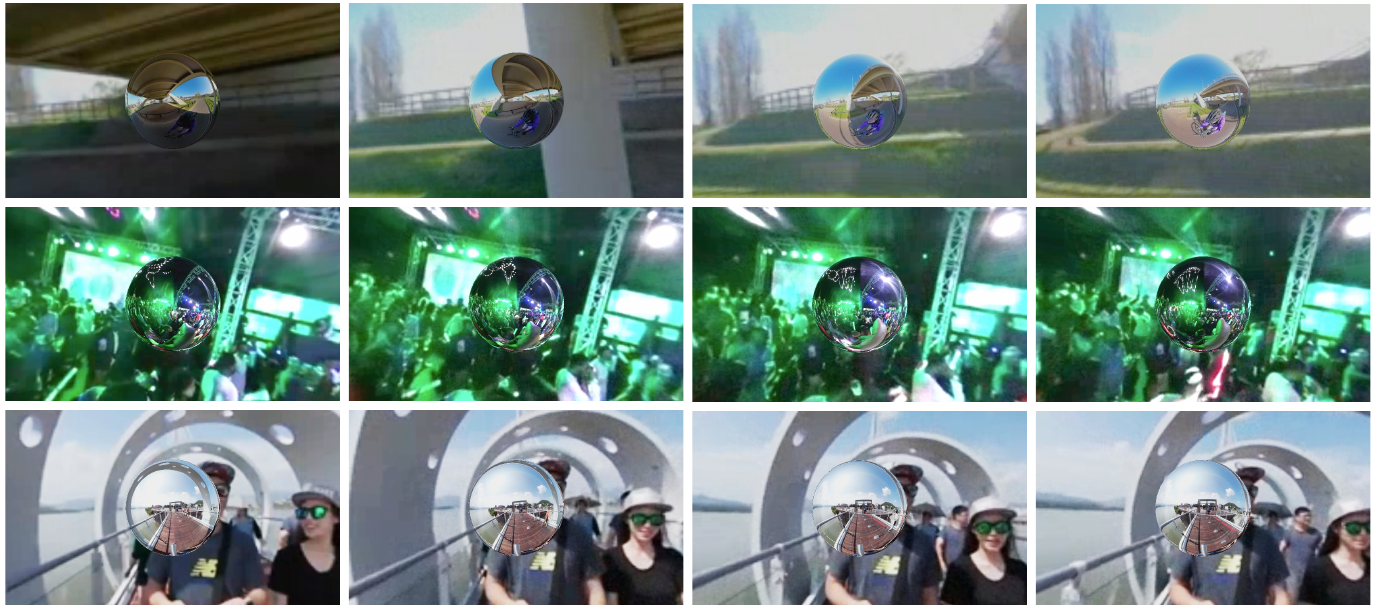}
    \caption{Samples from the V-LITESet dataset. Each row corresponds to a video, and the images in the row represent different timestamps sampled from the video.
    }
    \label{fig:dataset}
\end{figure}

\subsection{Generative Diffusion Priors}

Generative diffusion priors have been widely used in various computer vision tasks.
Due to their strong capability to model the complex distribution of natural images, they are naturally explored to handle low-level vision tasks (\eg, dehaze~\cite{dehaze}, colorization~\cite{lcad}, and super-resolution~\cite{sr}). This further motivates researchers to extend their application to classic vision tasks (\eg, object detection~\cite{detection}, segmentation~\cite{segmentation}, and pose estimation~\cite{pose}) and even 3D vision for reconstruction~\cite{reconstruction} and point cloud completion~\cite{pointcloud}.
Recently, researchers have also demonstrated that video diffusion models have zero-shot reasoning capabilities for the visual world, implicitly encoding complex physics, geometry, and material optics~\cite{videozeroshotleaner}. Based on these previous works, we reformulate the video lighting estimation problem and explore the approach to leverage the generative priors of video diffusion models to capture the dynamic illumination.

\section{V-LITESet Dataset}

Existing large-scale video generation datasets~\cite{bain2021webvid, wang2024internvid} primarily focus on LDR video content, making them ill-suited for developing HDR models. In contrast, existing HDR video datasets either focus on indoor scenarios~\cite{2017LavalIndoor} or provide limited in-the-wild samples~\cite{polyhaven2025}. To overcome these limitations, we introduce V-LITESet, a high-quality dataset specifically designed for training and evaluating HDR video lighting estimation.

\noindent\textbf{Dynamic spatiotemporal subset.}
We build the dynamic subset of V-LITESet upon the PanoVid dataset~\cite{xia2025panowan}, which includes diverse in-the-wild LDR panoramic videos. To lift these LDR videos to HDR, we employ a rigorous offline curation pipeline. Rather than assuming a single camera response function, we process the source panoramas using three distinct inverse tone-mapping curves (linear, $\gamma=2.2$, and filmic) to span a diverse set of candidate HDR distributions. To filter out severe photometric losses, we map these candidates back to the LDR domain, and query a VLM~\cite{qwen25vl} to evaluate the physical plausibility of the frames (\eg, unnatural color banding, severe overexposure, and irreversible detail loss). Approximately 57\% of the raw sequences are discarded due to low confidence scores. The remaining videos provide robust dynamic spatiotemporal context.

\noindent\textbf{Static photometric subset.}
Since photometric approximations for tonemapped HDR videos may introduce certain artifacts, we additionally integrate 800 high-quality HDR panoramic images from PolyHaven~\cite{polyhaven2025} to form the static subset of  V-LITESet. We adapt these images into static video sequences to match the input format of our video diffusion backbone. These real-world samples serve as absolute physical anchors during training, enabling the model to effectively learn the intrinsic luminance distributions and HDR contrasts of light sources (\eg, the accurate intensity ratio between the sun and the sky).

\noindent \textbf{Summary.}
The final V-LITESet comprises over 8K video pairs at $480 \times 832$ resolution with dynamic temporal lighting, alongside 800 HDR static sequences with diverse luminance distributions. This amounts to over 648K frames, split into 567K training and 81K evaluation samples. This provides a robust and comprehensive foundation for advancing research in dynamic HDR lighting estimation from a single video.

\section{Methodology}

\begin{figure*}[t]
    \centering
    \includegraphics[width=1\linewidth]{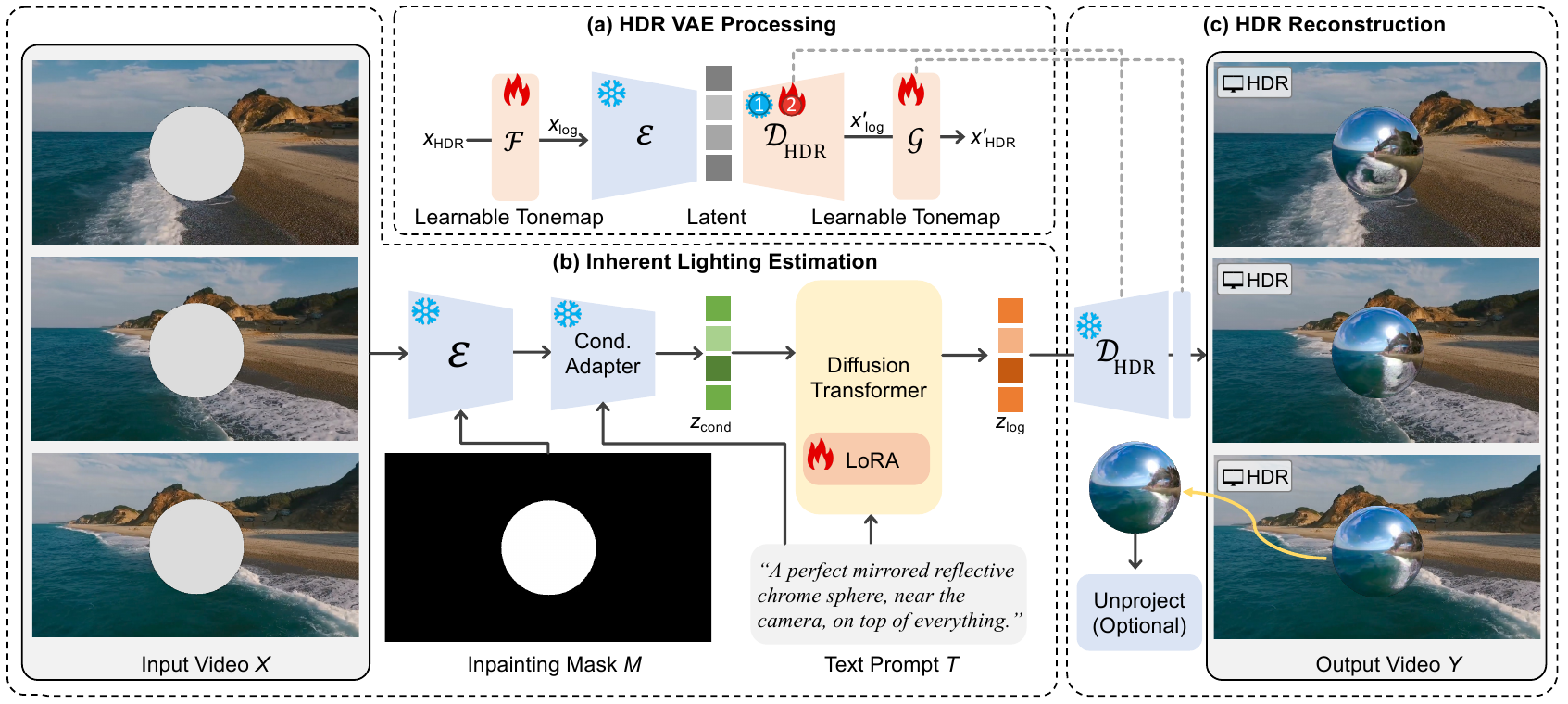}
    \caption{Overview of the V-LITE pipeline. \textbf{(a)} HDR VAE processing. To operate in log-domain space, an input HDR video $x_\text{HDR}$ is mapped by a learnable tone mapping network $\mathcal{F}$ to $x_\text{log}$, encoded by $\mathcal{E}$, and decoded by $\mathcal{D}_\text{HDR}$ into $x'_\text{log}$. An inverse network $\mathcal{G}$ then reconstructs the HDR video $x'_\text{HDR}$. We pretrain $\mathcal{F}$ and $\mathcal{G}$ before jointly optimizing them with $\mathcal{D}_\text{HDR}$. \textbf{(b)} Inherent lighting estimation. For an LDR video $X$ masked by $M$ (grey central region), $\mathcal{E}$ extracts latents that a condition adapter encodes into control signals $z_\text{cond}$. Conditioned on $z_\text{cond}$ and a text prompt $T$, a LoRA-adapted flow-matching diffusion Transformer inpaints the masked region, outputting the full latent sequence $z_\text{log}$. \textbf{(c)} HDR reconstruction. The HDR decoder $\mathcal{D}_\text{HDR}$ maps $z_\text{log}$ to the final video $Y$ featuring physically coherent reflections. Optionally, these chrome balls can be unprojected to estimate the scene's dynamic HDR environment map sequence, providing a powerful tool for downstream tasks like virtual object insertion.}
    \label{fig:pipeline}
\end{figure*}

In this section, we introduce the details of our V-LITE framework.
We first review the flow-matching video generation and editing models that form our foundation (\cref{sec:preliminary}).
Next, we propose the end-to-end VAE for HDR video processing to adapt the pretrained video backbone into the HDR domain (\cref{sec:vae}).
Finally, we present our strategy to reformulate lighting estimation as a video inpainting task, leveraging its inherent generative priors to accurately capture the scene’s dynamic lighting (\cref{sec:latent}). The overall pipeline is shown in \cref{fig:pipeline}.

\subsection{Preliminaries} \label{sec:preliminary}

\noindent\textbf{Video generation models}
Video generation models have demonstrated strong capabilities in producing diverse and high-quality dynamic content. As a representative practice of the video backbone, Wan~2.1~\cite{wan2025wan} adopts a diffusion Transformer architecture, modeling complex spatio-temporal dynamics through a latent-space training scheme based on the flow matching framework~\cite{lipman2023fm}.

Given a target video $x_1$, the pretrained VAE encoder $\mathcal{E}$ first maps it into a latent code $z_1 = \mathcal{E}(x_1)$.
During training, the model randomly samples a noise latent $z_{0} \sim \mathcal{N}(\mathbf{0}, \mathbf{I})$, and creates an intermediate latent code via linear interpolation for a timestep $t \in [0,1]$: $z_t = t \cdot z_1 + (1 - t) \cdot z_0$.
The model is trained to predict the ground-truth velocity field associated with the interpolation path. Formally, the training objective is:
\begin{align}
\mathcal{L} = \mathbb{E}_{z_0, z_1, c_{\mathrm{txt}}, t} \big [ \| v_\theta(z_t, c_{\mathrm{txt}}, t) - (z_1 - z_0) \|^2 \big ],
\end{align}
where $c_{\mathrm{txt}}$ denotes the text condition and $\theta$ represents the model parameters. $v_\theta(\cdot)$ is the diffusion Transformer to predict the velocity field at latent code $z_t$.

\noindent\textbf{Video editing models}
Video editing aims to modify the video content based on the text description indicating the editing direction and the semantics of the original video.
To selectively transform targeted regions while maintaining temporal and spatial coherence in unedited areas, a recent method~\cite{jiang2025vace} proposes to incorporate text, mask, and other visual conditions into a unified conditioning representation. This conditioning representation is injected into  the video backbone via an additional adapter. The training objective is thus formulated as:
\begin{align} \label{eq:editing}
\mathcal{L} = \mathbb{E}_{z_0, z_1, \mathcal{V}, t}
\big[ \| v_\theta(z_t, \mathcal{V}, t) - (z_1 - z_0) \|^2 \big],
\end{align}
where $\mathcal{V} = (T, V, M)$ denotes the conditioning input, with $T$ representing the text description, $V$ denoting the visual conditions, and $M$ providing the corresponding masks.

\subsection{HDR VAE Processing} \label{sec:vae}

The success of leveraging inherent generative priors for lighting estimation depends critically on operating in the HDR domain. However, since existing video generation backbones~\cite{yang2024cogvideox, wan2025wan} are designed for LDR content, it is essential to adapt their architectures to accurately support end-to-end HDR processing.

Existing methods~\cite{phongthawee2024diffusionlight,chinchuthakun2025diffusionlightturbo,liang2025luxdit} typically address this gap by producing multiple LDR variants (\eg, exposure-bracketed frames) that are subsequently fused to approximate an HDR result.
However, such approaches require multiple inference passes or an additional fusion stage, leading to increased computational cost and potential error accumulation across the reconstruction pipeline.
In contrast, we propose to integrate the HDR domain conversion directly into the VAE, bridging the distribution gap within the latent space.
As presented in~\cref{fig:pipeline}~(a), this design inherently makes the LDR-trained video backbone compatible with HDR data.

\noindent\textbf{Tonemap adapters.}
We first introduce a pair of learnable tonemap adapters and integrate them into the VAE~\cite{wan2025wan} for end-to-end HDR video processing.
Specifically, given a linear HDR video $x_\mathrm{HDR}$, our tonemap adapter converts it to the log domain. It then applies a single 3D convolutional layer $\mathcal{F}$, a learnable scale $\mathbf{s}$ and bias $\mathbf{b}$ to calculate the log-domain representation:
\begin{align}\label{eq:tonemap}
x_{\log} = \mathbf{s} \odot \left[\log(x_\mathrm{HDR}+\epsilon) + \mathcal{F}\big(\log(x_\mathrm{HDR}+\epsilon)\big)\right] + \mathbf{b}.
\end{align}
This adaptive mapping enables the encoder to dynamically stretch and compress signal ranges. The log domain representation $x_{\log}$ is fed into the LDR VAE encoder $\mathcal{E}$ to generate the compressed latent code $z=\mathcal{E}(x_{\log})$.
After that, the HDR VAE decoder $\mathcal{D}_\mathrm{HDR}$ produces the log-domain representation $x'_{\log} = \mathcal{D}_\mathrm{HDR}(z)$. The inverse tonemap adapter then maps this representation back to the linear HDR domain:
\begin{align}\label{eq:inverse_remap}
x'_{\mathrm{HDR}}
= \exp\!\left[
\frac{x'_{\log} - \mathbf{b}}{\mathbf{s}}
+ \mathcal{G}\!\left( \frac{x'_{\log} - \mathbf{b}}{\mathbf{s}} \right)
\right],
\end{align}

where $\mathcal{G}$ is a single 3D convolutional layer and $x'_{\mathrm{HDR}}$ is the reconstructed HDR video. During the training process, these lightweight adapters learn the dynamic illumination and extreme ranges within a stable log-domain latent space. Notably, our HDR VAE decoder has the exact same architecture as the pretrained LDR VAE from the video backbone~\cite{wan2025wan}.

\noindent\textbf{Training scheme.}
The VAE training is formulated as a reconstruction task, requiring the model to minimize the error between the input video $x_{\mathrm{HDR}}$ and the reconstructed video $x'_{\mathrm{HDR}}$.
To adapt our VAE with tonemap adapters for stable HDR processing, we employ a two-stage finetuning strategy.
In the first stage, we freeze all VAE parameters and train the two tonemap adapters to establish a stable log-domain representation.
In the second stage, we unfreeze the VAE decoder and jointly finetune it with the adapters.
After finetuning, the combined VAE and tonemap adapters operate directly in the HDR domain, preserving visual fidelity across extreme luminance ranges typical of HDR data.

\subsection{Inherent Lighting Estimation} \label{sec:latent}
Previous work~\cite{liang2025luxdit} employs video generation models as a video translator for lighting estimation.
In contrast, observing that recent video generation models~\cite{wan2025wan} demonstrate an implicit ability to model lighting, producing temporally and spatially light-coherent results, we are motivated to directly exploit their internal lighting modeling capability rather than formulating the task as a translation problem.
As illustrated in \cref{fig:pipeline}~(b), we reformulate the task of video lighting estimation as an inpainting problem.

Since the central region of the scene is typically irrelevant for global illumination, we leverage a central mask $M$ to define the inpainting region for the synthesis of a physically plausible chrome ball.
Given an in-the-wild LDR video $x_{\mathrm{LDR}}$, we construct the masked input as $x_{\mathrm{input}} = x_{\mathrm{LDR}} \odot (1 - M) + C \odot M$, where $C$ is a constant gray value.
We then decouple this input into the environment region $x_e = x_{\mathrm{input}} \odot M$ and the scene content $x_s = x_{\mathrm{input}} \odot (1 - M)$ as two conditions for the generation model.
This setup allows the model to simultaneously reconstruct the HDR environment map from the surrounding regions and enhance the dynamic range of the original scene.

\noindent\textbf{Diffusion process.}
During training and inference, both $x_s$ and $x_e$ are represented in the latent space as $z_s$ and $z_e$ using the pretrained LDR VAE encoder $\mathcal{E}$. At each inference step $t$, we utilize a condition adapter~\cite{jiang2025vace} to inject these latent codes into the video backbone to predict the velocity field:
\begin{align}
    \hat{v} = f_{\theta}\bigl(\hat{z}_t, f_{c}([z_s, z_e, c_{\mathrm{txt}}]), c_{\mathrm{txt}})\bigr),
\end{align}
where $f_{\theta}$ denotes the diffusion Transformer of the video backbone, $f_{c}$ denotes the condition adapter for video inpainting, and $c_{\mathrm{txt}}$ represents the text embedding driving the model to fill the chrome ball.
During inference, we iteratively solve the ordinary differential equation $\textrm{d}z_t/\textrm{d}t = \hat{v}$ from $t = 1$ to $t = 0$ using a numerical solver to synthesize the chrome ball.
In particular, the prompt we provide is \textit{``A perfect mirrored reflective chrome sphere, near the camera, on top of everything''}, which explicitly specifies the appearance and placement of the chrome ball.

After the diffusion process, our pretrained VAE decoder $\mathcal{D}_\mathrm{HDR}$ is employed to reconstruct HDR video frames $\hat{x}_{\mathrm{HDR}}$. To extract the final HDR environment map $\mathbf{E}$, we perform a post-processing step, where the chrome ball is isolated from the reconstructed HDR frames using the corresponding mask and re-projected into the environment map:
\begin{align}
\mathbf{E} = f_{\mathcal{R}}(\hat{x}_{\mathrm{HDR}} \odot M),
\end{align}
where $f_{\mathcal{R}}$ denotes the equirectangular transformation. This procedure produces a high-fidelity and temporally coherent sequence of HDR environment maps that faithfully capture the dynamic lighting of the scene.

\noindent\textbf{Optimization process.}
We train only the core video diffusion Transformer using LoRA-based fine-tuning~\cite{hu2021lora} to align its generative priors with the HDR latent distribution.
During training, we randomly mix samples from the dynamic spatiotemporal and static photometric subsets at a $10:1$ ratio. To enable this joint training, we replicate the HDR images within the photometric subset to form static video sequences. Consequently, this mixed-data training strategy ensures that the model preserves its inherent temporal consistency priors through the dynamic videos, while utilizing the static sequences as physical anchors to learn accurate luminance distributions (\eg, precise sun intensities).

To simultaneously reconstruct the HDR environment map and enhance the dynamic range of the original scene context, we define our objective as follows:
\begin{align}
\mathcal{L} &= \mathbb{E}_{z, \epsilon, t}
\big[ \|M' \odot (v_\theta(z_t, \mathcal{C}, t) - v )\|^2 \big],
\end{align}
where $\mathcal{C}=[z_\text{cond}, c_{\mathrm{txt}}]$ denotes the full conditioning input, $z_\text{cond} = \{z_e,z_s\}$, and $\epsilon \sim \mathcal{N}(\mathbf{0}, \mathbf{I})$ represents a latent noise sample. Here, $v=z-\epsilon$ denotes the target velocity, $v_\theta$ is the predicted velocity field, and $M'=1+(\alpha-1)M$ is a modulation mask with a weighting factor $\alpha=5$ to explicitly emphasize the lighting estimation within the central inpainting region.

\section{Experiments}
In this section, we conduct a series of experiments to validate the effectiveness of our proposed method, V-LITE. We first introduce the experimental setup, including our implementation details, the dataset used, and the evaluation metrics. We then present comprehensive quantitative and qualitative comparisons against state-of-the-art methods. Finally, we perform detailed ablation studies to analyze the contribution of each key component in our framework.

We compare V-LITE against DiffusionLight~\cite{phongthawee2024diffusionlight}, DiffusionLightTurbo~\cite{chinchuthakun2025diffusionlightturbo} and StyleLight~\cite{wang2022stylelight}. Since most existing methods do not support dynamic lighting from a single video, we adapt state-of-the-art single-image lighting estimation methods to the video domain by applying them in a per-frame manner.

\subsection{Implementation Details}

Our framework is built upon the pretrained Wan~2.1-1.3B video model~\cite{wan2025wan}. Our custom-trained HDR-aware VAE is specifically designed to operate in the linear Rec. 2020 color space, which provides a wide-gamut, scene-linear representation aligned with HDR standards, ensuring physically accurate color reproduction. To effectively model the high dynamic range, the VAE's latent codes are compressed into a log space before being processed by the core diffusion model. For HDR-aware VAE fine-tuning, we adopt a two-stage training scheme with $10$k steps in the first stage and $5$k steps in the second. We optimize the model using the standard VAE objective, which includes the reconstruction loss and KL divergence loss, identical to the losses used in the original VAE training. For diffusion Transformer fine-tuning, we employ LoRA~\cite{hu2021lora} with a rank of $r=32$ and apply it to the attention layers of the diffusion Transformer. The model is trained on our V-LITESet for 100K steps with a global batch size of 64 on 8 NVIDIA H100 GPUs. We use the AdamW optimizer with a learning rate of $1 \times 10^{-5}$. During training and inference, the input videos containing the masked sphere are processed at a resolution of $480 \times 832$. After the model inpaints the sphere, its reflection is unwrapped to produce the final HDR environment map at a resolution of $256 \times 512$, which is commonly used in prior work~\cite{phongthawee2024diffusionlight,chinchuthakun2025diffusionlightturbo,liang2025luxdit,wang2022stylelight}.

\subsection{Quantitative Evaluation}

\begin{table*}[tb]
\centering
\caption{Quantitative Comparison. We highlight the best score in boldface. $^\dagger$Comparison on a subset of 10 videos.}
\label{tab:main_quantitative_results}
\setlength\tabcolsep{8pt}
\resizebox{\linewidth}{!}{
\begin{tabular}{l|cccc|cc|c}
\toprule
\multicolumn{1}{c|}{} &
\multicolumn{4}{c|}{\textbf{Editable Indoor}} &
\multicolumn{2}{c|}{\textbf{EnvMapNet}} &
\textbf{Efficiency}  \\
\cmidrule(r){2-5} \cmidrule(lr){6-7} \cmidrule(l){8-8}
\textbf{Method} &
MSE $\downarrow$ & SI-MSE $\downarrow$ &
AER $\downarrow$ & LS $\downarrow$ &
AED $\downarrow$ & AS $\downarrow$ & Seconds $\downarrow$
\\
\midrule
DiffusionLight$^\dagger$~\cite{phongthawee2024diffusionlight}
& 0.10 & 0.05 & \textbf{4.58} & \textbf{0.03} & \textbf{40.07} & 16.76 & 145{,}800 \\
Ours$^\dagger$
& \textbf{0.09} & \textbf{0.03} & 4.68 & \textbf{0.03} & 42.53 & \textbf{15.96} & \textbf{80} \\
\midrule
DiffusionLightTurbo~\cite{chinchuthakun2025diffusionlightturbo}
& 0.11 & 0.05 & 4.90 & 0.04 & \textbf{37.74} & 16.68 & 713 \\
StyleLight~\cite{wang2022stylelight}
& 0.13 & 0.07 & 6.01 & 0.05 & 44.22 & 17.83 & 2{,}002 \\
Ours &
\textbf{0.10} & \textbf{0.03} & \textbf{4.77} & \textbf{0.03} &
41.63 & \textbf{16.30} & \textbf{80} \\
\midrule
LDR Baseline &
0.11 & 0.04 & 5.45 & 0.03 &
44.71 & 15.87 & 80 \\
Frozen backbone &
3.00 & 0.07 & 12.95 & 0.04 &
51.71 & 19.45 & 80\\
Video-only & 0.10 & 0.03 & 4.82 & 0.02 & 41.96 & 15.81 & 80 \\
\bottomrule
\end{tabular}
}
\end{table*}

\noindent\textbf{Evaluation protocol.}
We conduct quantitative evaluations on two public benchmarks following the evaluation protocol of DiffusionLight~\cite{phongthawee2024diffusionlight}.
\textit{(i)} \textit{Editable Indoor} assesses illumination quality through scene rendering and reports MSE, SI-MSE, Angular Error in Radians (AER), and Lighting Stability (LS, the standard deviation of SI-MSE).
\textit{(ii)} \textit{EnvMapNet} estimates the dominant lighting direction from the environment maps and reports both the Angular Error in Degrees (AED) and Angle Stability (AS, the standard deviation of AED).

\noindent\textbf{Baselines.}
We compare our method against DiffusionLight~\cite{phongthawee2024diffusionlight}, DiffusionLight-Turbo~\cite{chinchuthakun2025diffusionlightturbo}, and StyleLight~\cite{wang2022stylelight}. Note that DiffusionLight is extremely time-consuming (\ie, over one day for a single video), so we report comparisons on a subset of 10 videos. The quantitative results are provided in \cref{tab:main_quantitative_results}. Compared to DiffusionLight, our model achieves competitive lighting accuracy and superior temporal stability while requiring only a fraction of the computation time. Compared to DiffusionLight-Turbo and StyleLight, our model attains the best overall performance.

\begin{figure*}[!t]

\def\lh{0.88cm}
\def\ftsz{\footnotesize}

\renewcommand\tabcolsep{0.0pt}

\renewcommand{\arraystretch}{0}

\newcommand{\rinput}[1]{\hspace{1pt}\includegraphics[height=\lh]{#1}}

\newcommand{\rowtop}{\rule{0pt}{0.77cm}}

\centering \small
\begin{tabular}{cccccc@{\hspace{2pt}}:@{\hspace{2pt}}cccccc}
\rowtop\includegraphics[height=\lh]{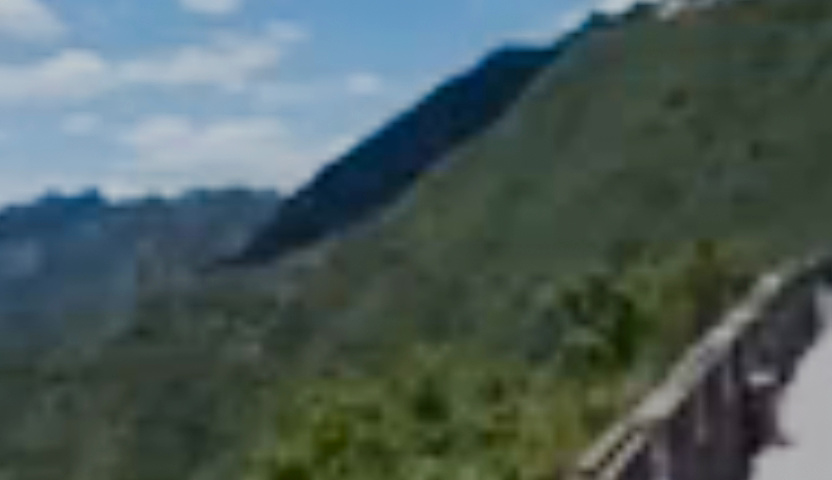} &
\includegraphics[height=\lh]{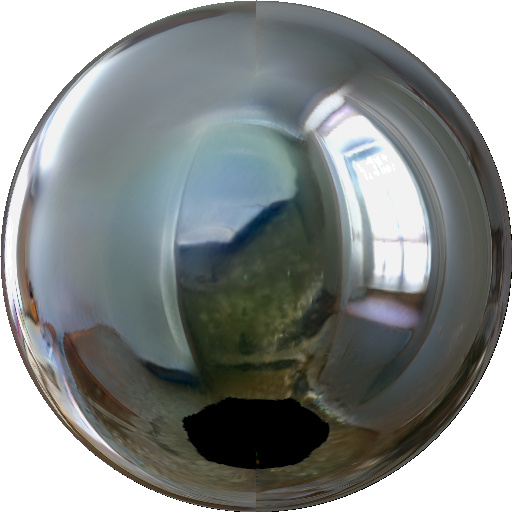} &
\includegraphics[height=\lh]{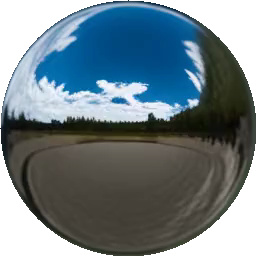} &
\includegraphics[height=\lh]{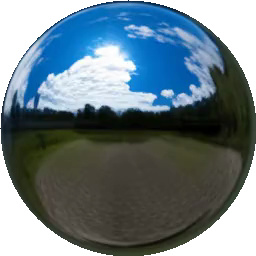} &
\includegraphics[height=\lh]{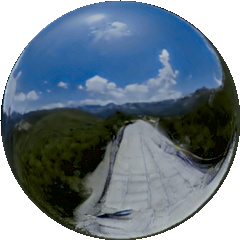} &
\includegraphics[height=\lh]{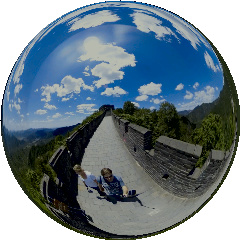} &
\rinput{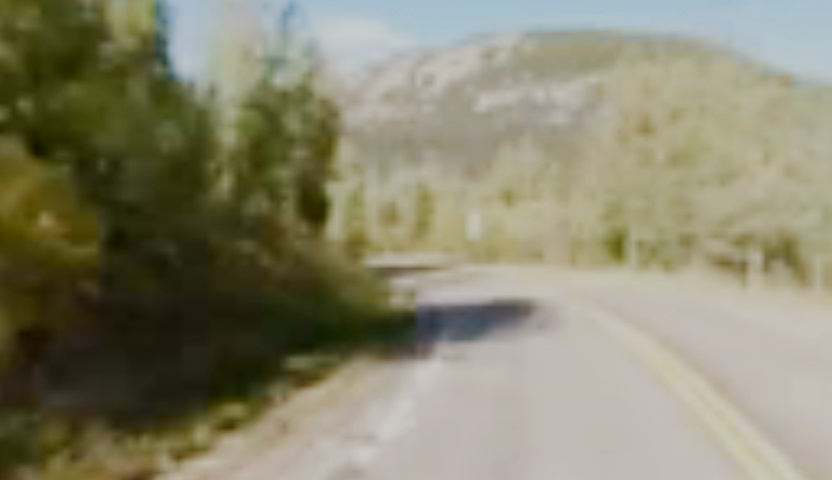} &
\includegraphics[height=\lh]{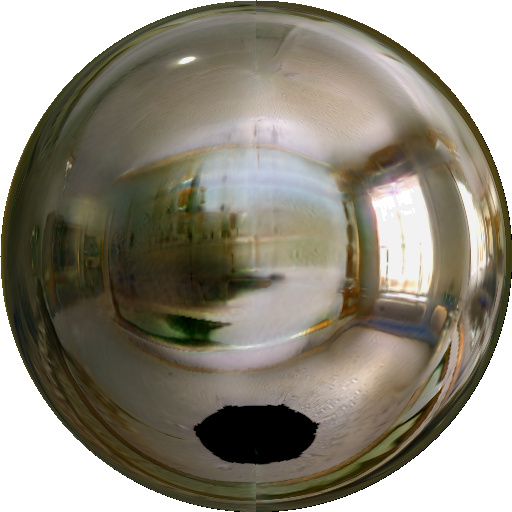} &
\includegraphics[height=\lh]{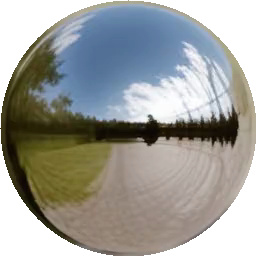} &
\includegraphics[height=\lh]{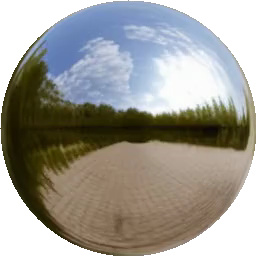} &
\includegraphics[height=\lh]{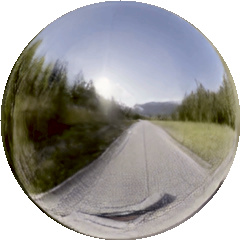} &
\includegraphics[height=\lh]{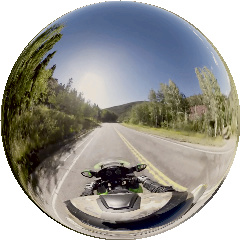} \\
\includegraphics[height=\lh]{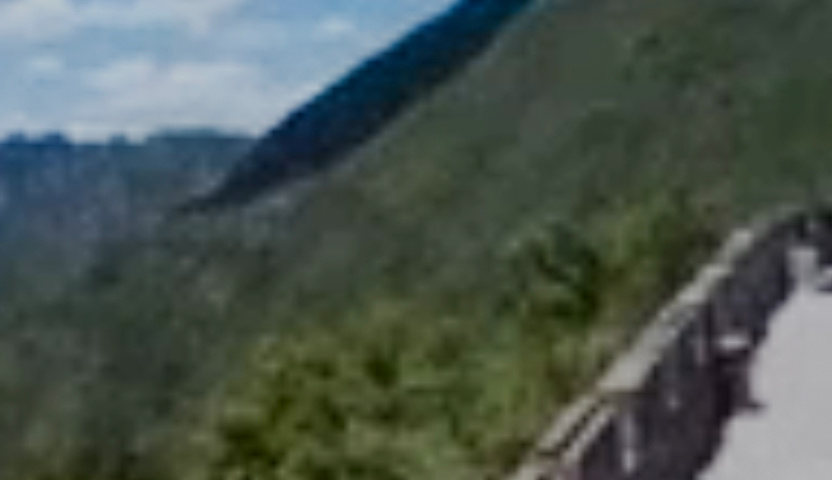} &
\includegraphics[height=\lh]{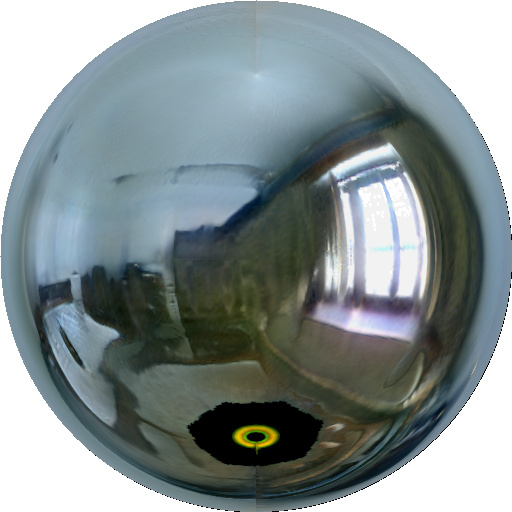} &
\includegraphics[height=\lh]{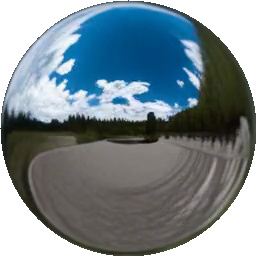} &
\includegraphics[height=\lh]{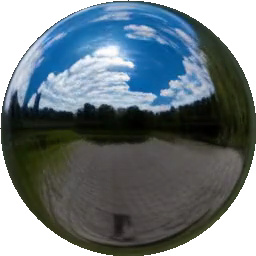} &
\includegraphics[height=\lh]{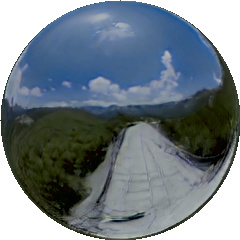} &
\includegraphics[height=\lh]{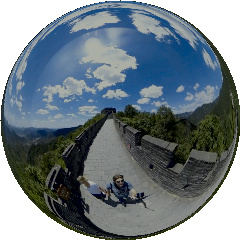} &
\rinput{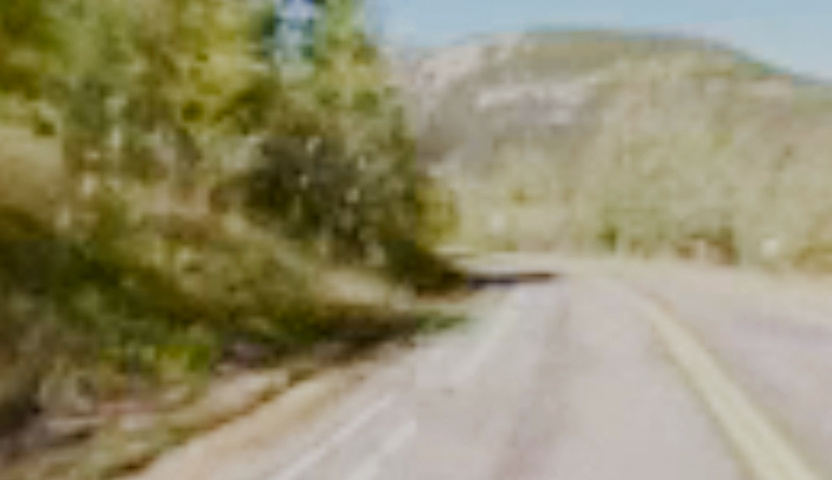} &
\includegraphics[height=\lh]{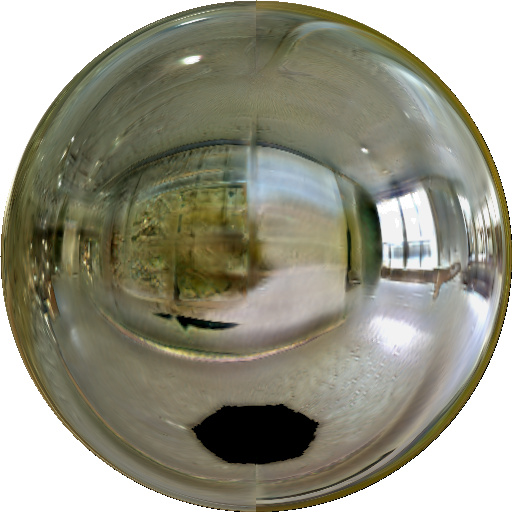} &
\includegraphics[height=\lh]{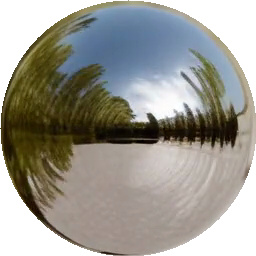} &
\includegraphics[height=\lh]{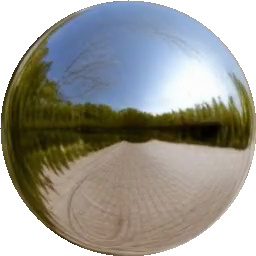} &
\includegraphics[height=\lh]{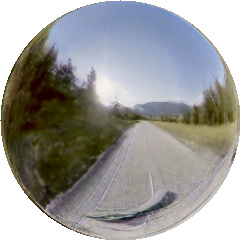} &
\includegraphics[height=\lh]{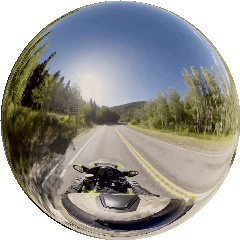} \\
\includegraphics[height=\lh]{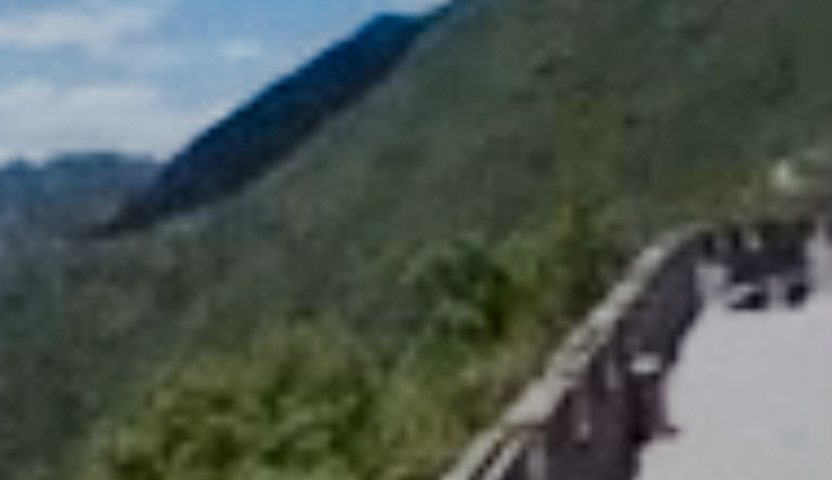} &
\includegraphics[height=\lh]{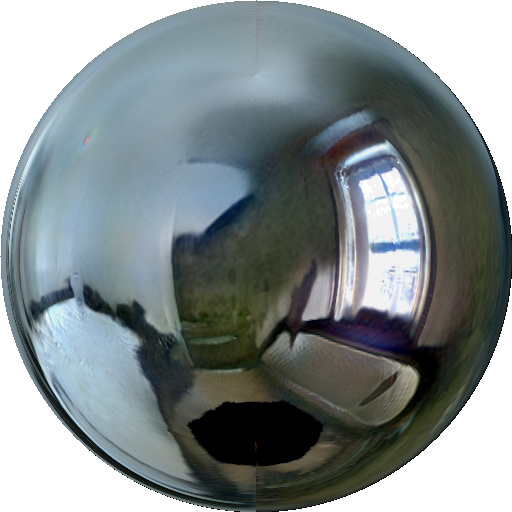} &
\includegraphics[height=\lh]{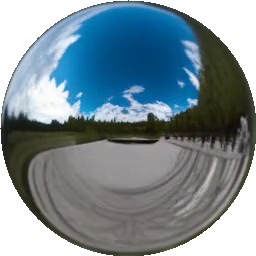} &
\includegraphics[height=\lh]{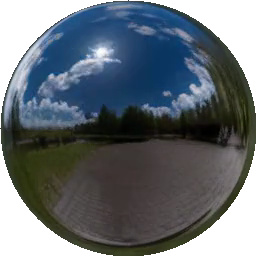} &
\includegraphics[height=\lh]{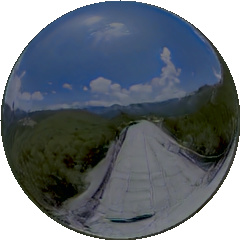} &
\includegraphics[height=\lh]{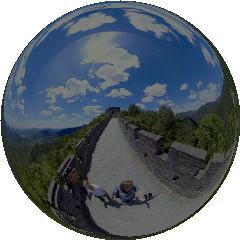} &
\rinput{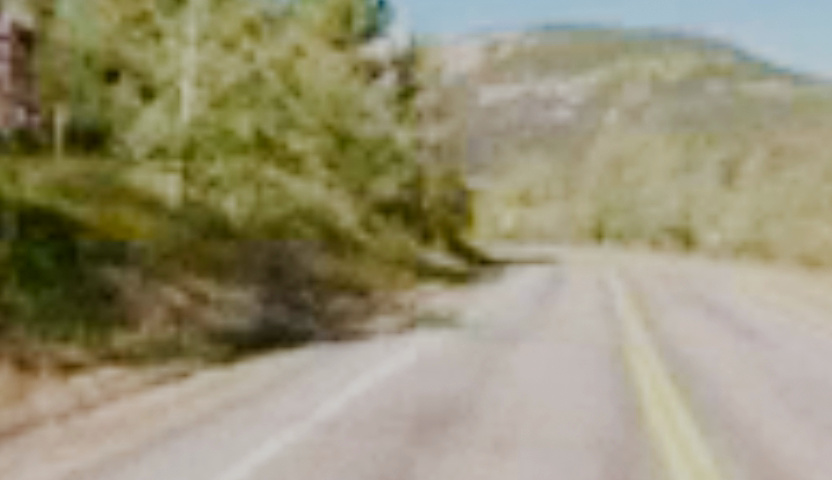} &
\includegraphics[height=\lh]{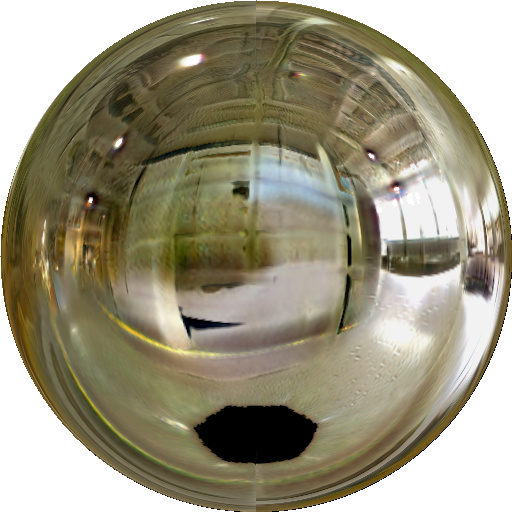} &
\includegraphics[height=\lh]{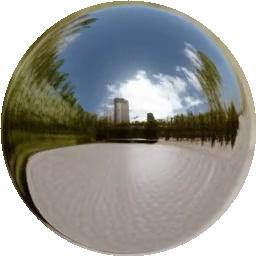} &
\includegraphics[height=\lh]{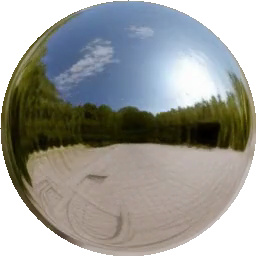} &
\includegraphics[height=\lh]{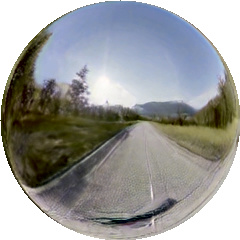} &
\includegraphics[height=\lh]{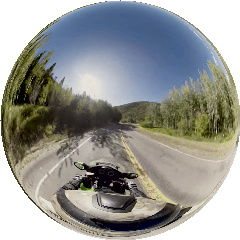} \\
\includegraphics[height=\lh]{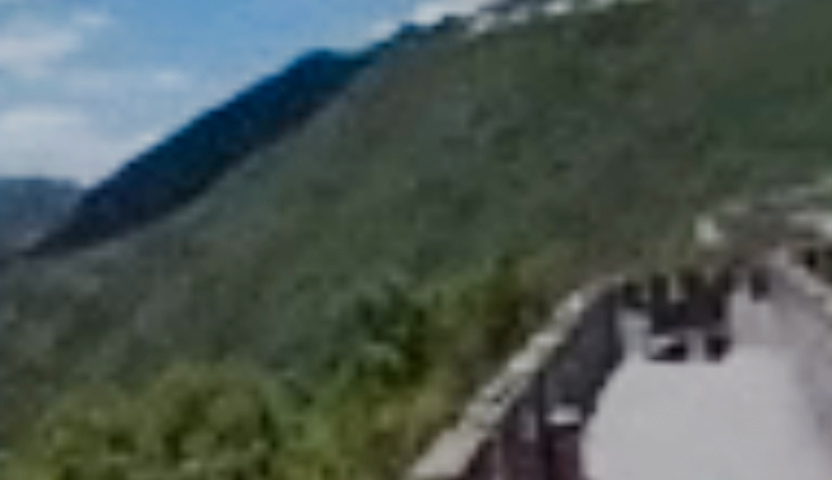} &
\includegraphics[height=\lh]{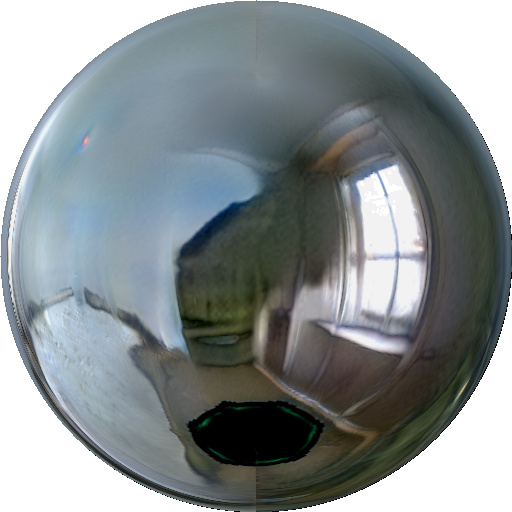} &
\includegraphics[height=\lh]{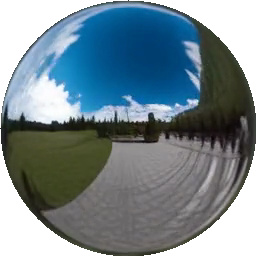} &
\includegraphics[height=\lh]{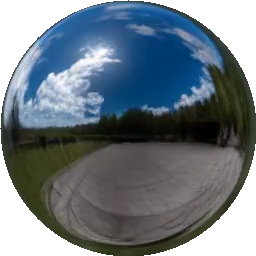} &
\includegraphics[height=\lh]{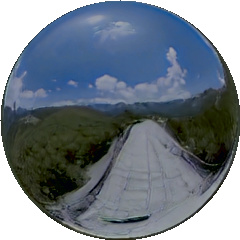} &
\includegraphics[height=\lh]{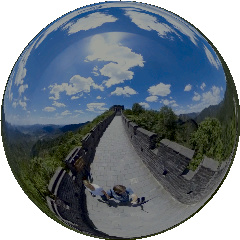} &
\rinput{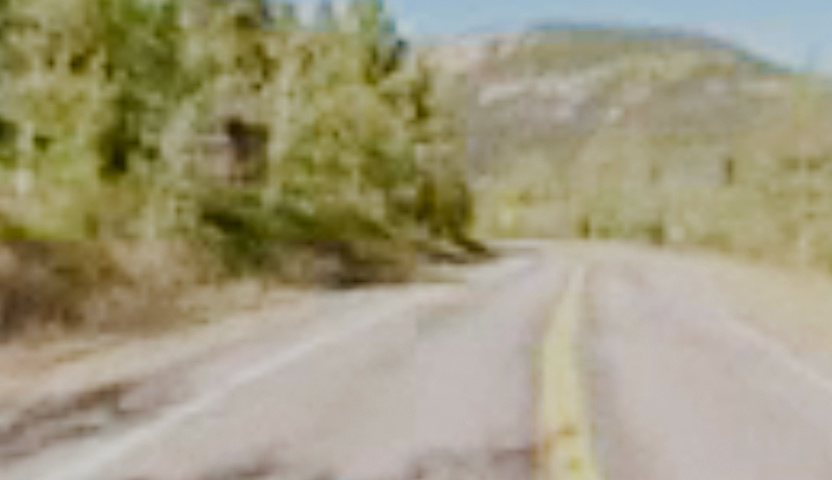} &
\includegraphics[height=\lh]{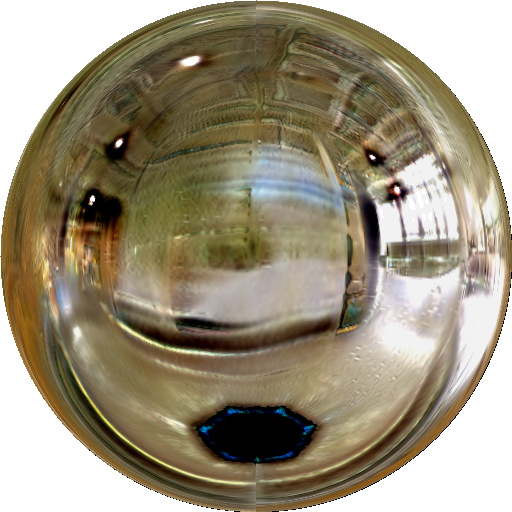} &
\includegraphics[height=\lh]{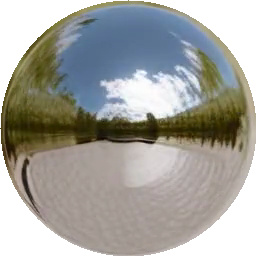} &
\includegraphics[height=\lh]{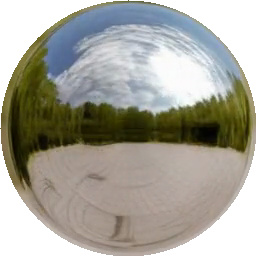} &
\includegraphics[height=\lh]{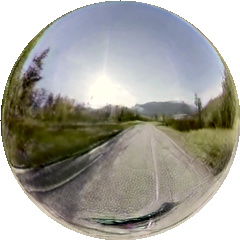} &
\includegraphics[height=\lh]{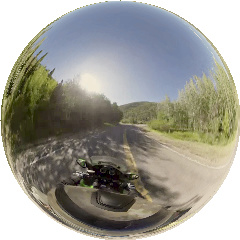} \\

\hdashline

\includegraphics[height=\lh]{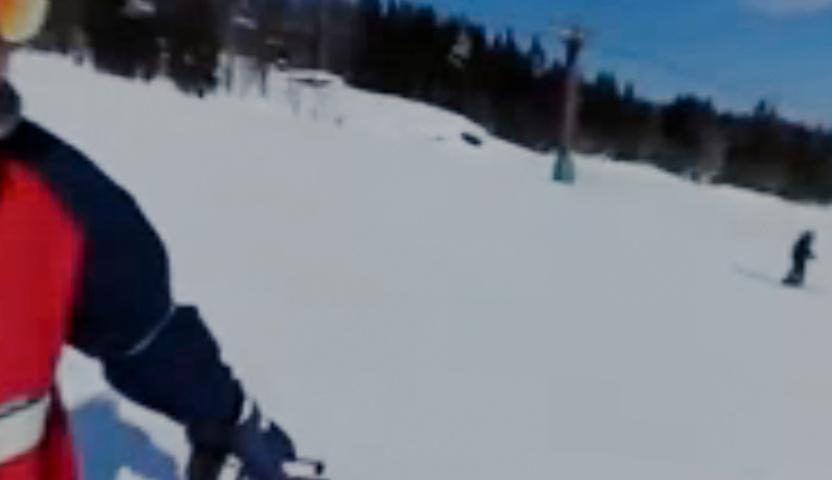} &
\includegraphics[height=\lh]{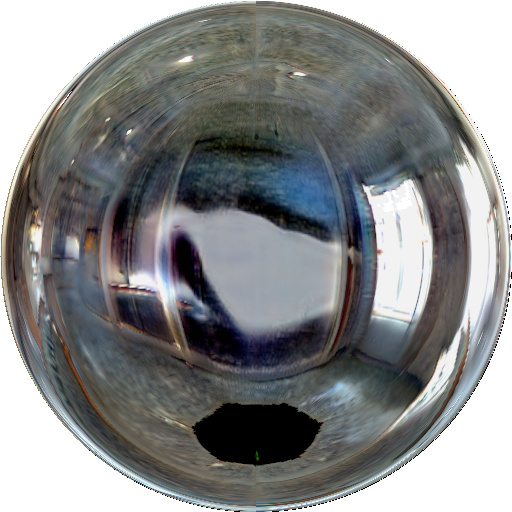} &
\includegraphics[height=\lh]{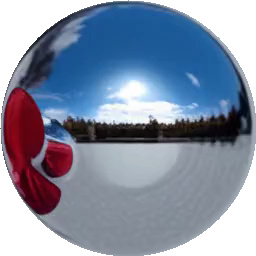} &
\includegraphics[height=\lh]{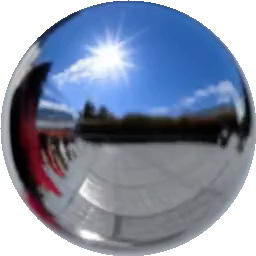} &
\includegraphics[height=\lh]{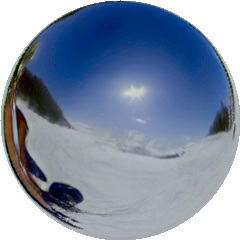} &
\includegraphics[height=\lh]{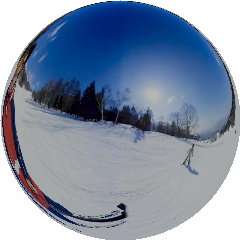} &
\rinput{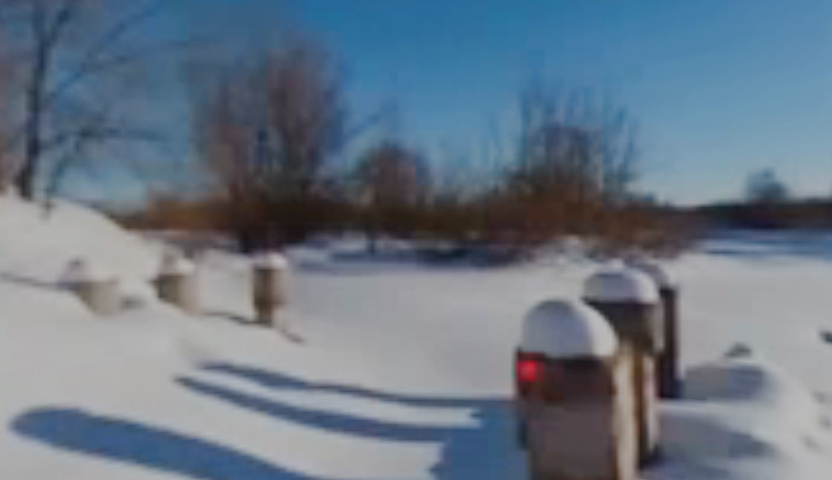} &
\includegraphics[height=\lh]{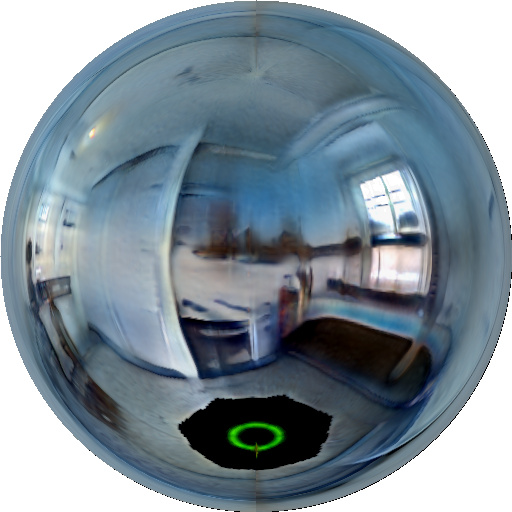} &
\includegraphics[height=\lh]{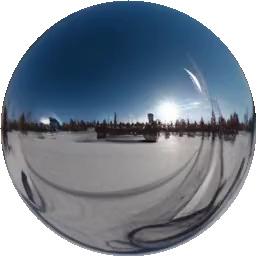} &
\includegraphics[height=\lh]{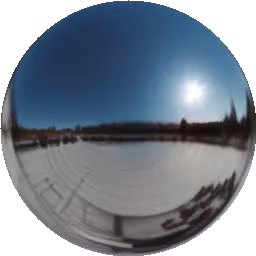} &
\includegraphics[height=\lh]{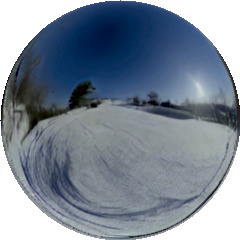} &
\includegraphics[height=\lh]{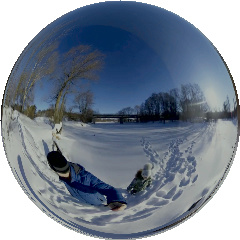} \\
\includegraphics[height=\lh]{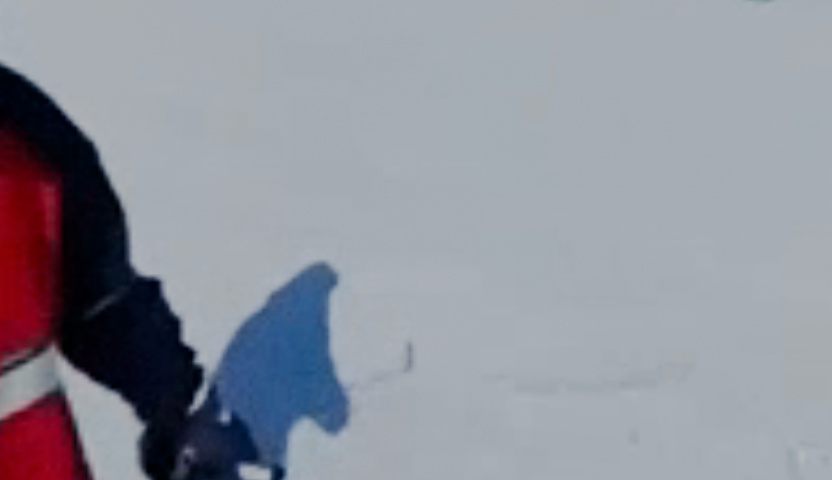} &
\includegraphics[height=\lh]{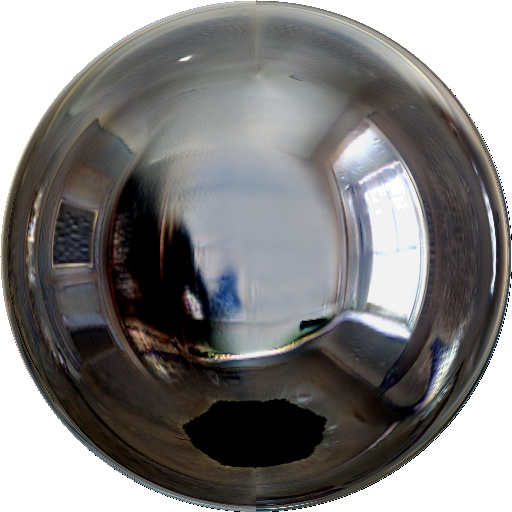} &
\includegraphics[height=\lh]{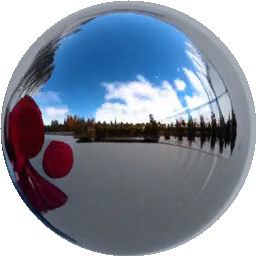} &
\includegraphics[height=\lh]{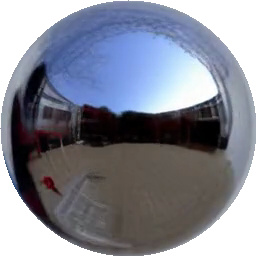} &
\includegraphics[height=\lh]{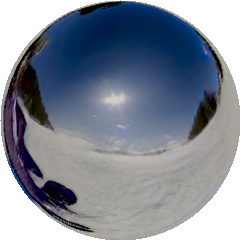} &
\includegraphics[height=\lh]{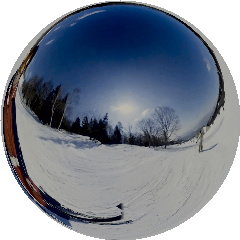} &
\rinput{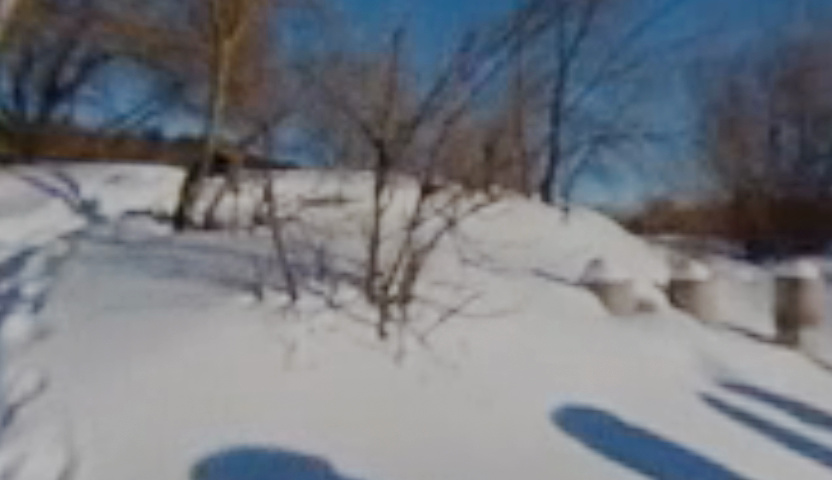} &
\includegraphics[height=\lh]{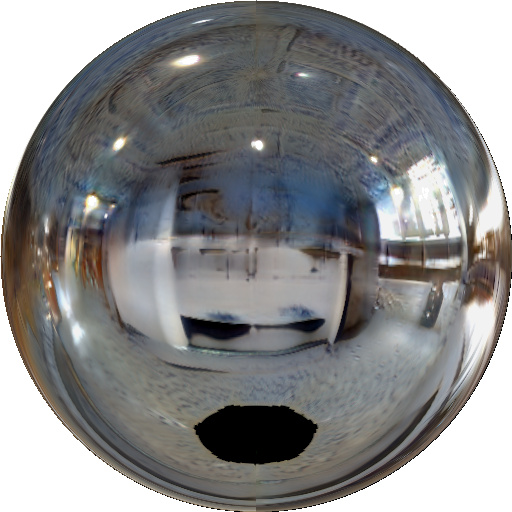} &
\includegraphics[height=\lh]{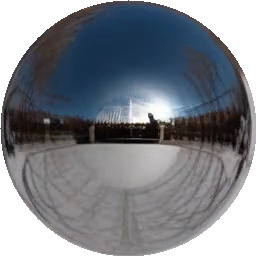} &
\includegraphics[height=\lh]{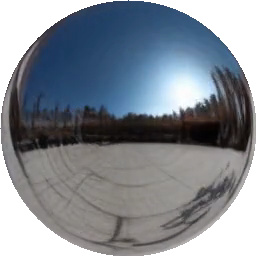} &
\includegraphics[height=\lh]{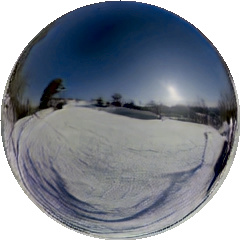} &
\includegraphics[height=\lh]{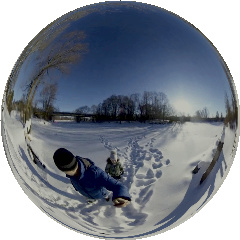} \\
\includegraphics[height=\lh]{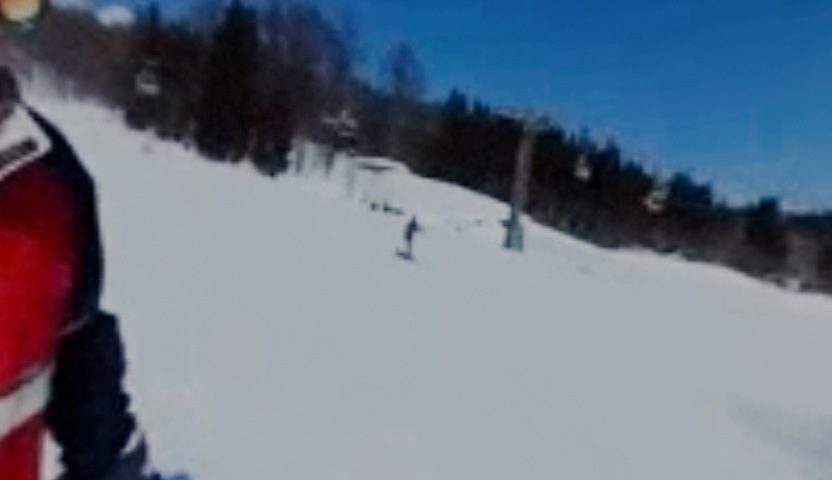} &
\includegraphics[height=\lh]{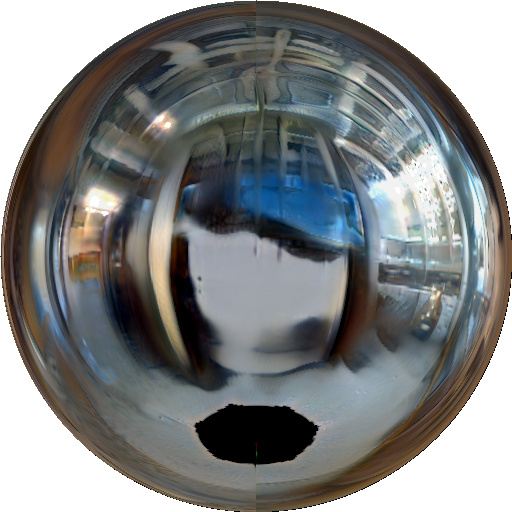} &
\includegraphics[height=\lh]{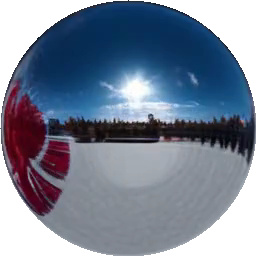} &
\includegraphics[height=\lh]{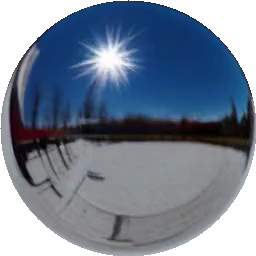} &
\includegraphics[height=\lh]{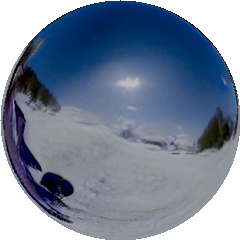} &
\includegraphics[height=\lh]{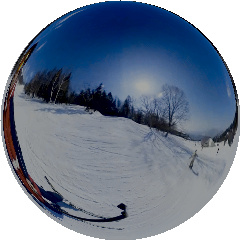} &
\rinput{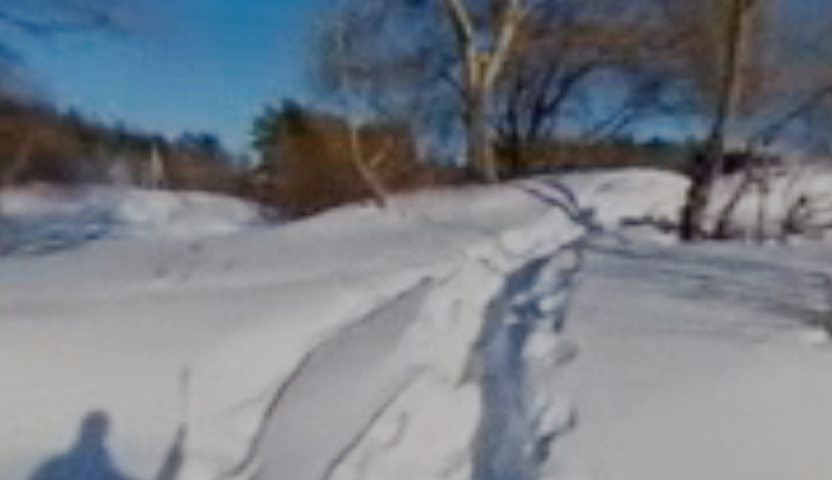} &
\includegraphics[height=\lh]{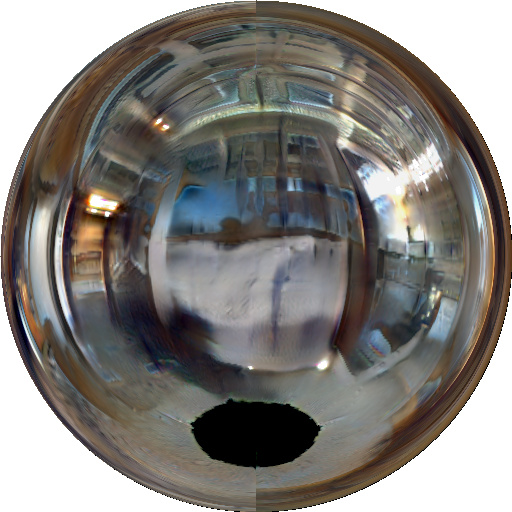} &
\includegraphics[height=\lh]{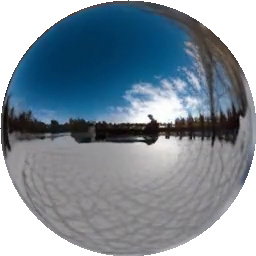} &
\includegraphics[height=\lh]{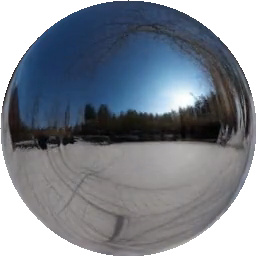} &
\includegraphics[height=\lh]{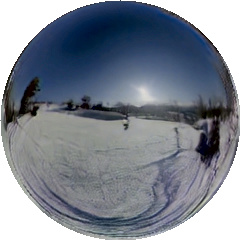} &
\includegraphics[height=\lh]{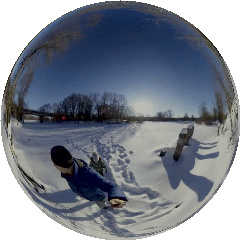} \\
\includegraphics[height=\lh]{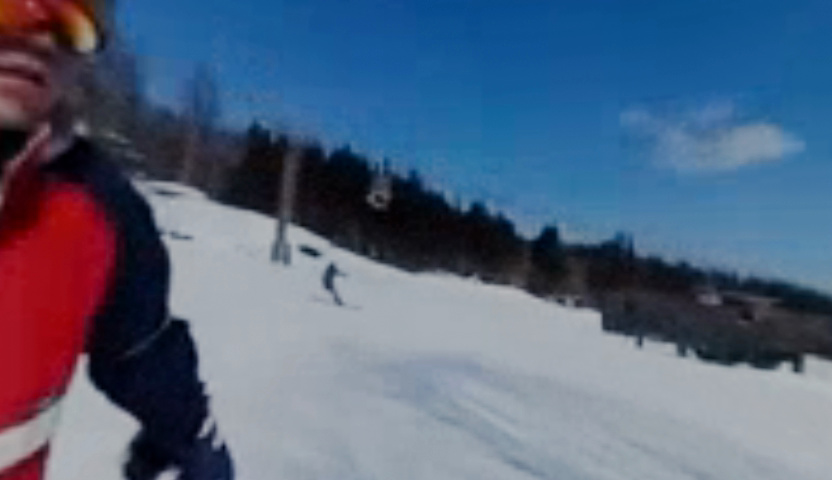} &
\includegraphics[height=\lh]{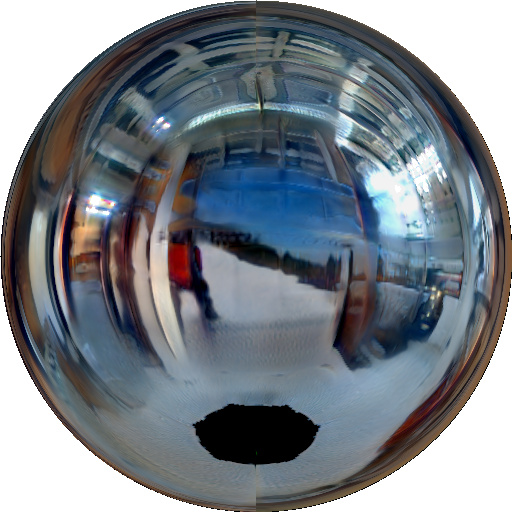} &
\includegraphics[height=\lh]{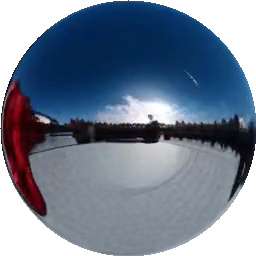} &
\includegraphics[height=\lh]{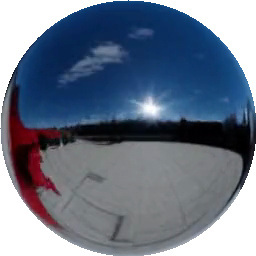} &
\includegraphics[height=\lh]{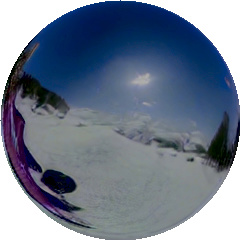} &
\includegraphics[height=\lh]{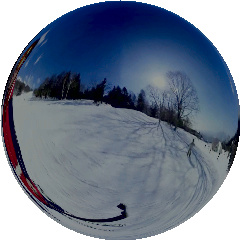} &
\rinput{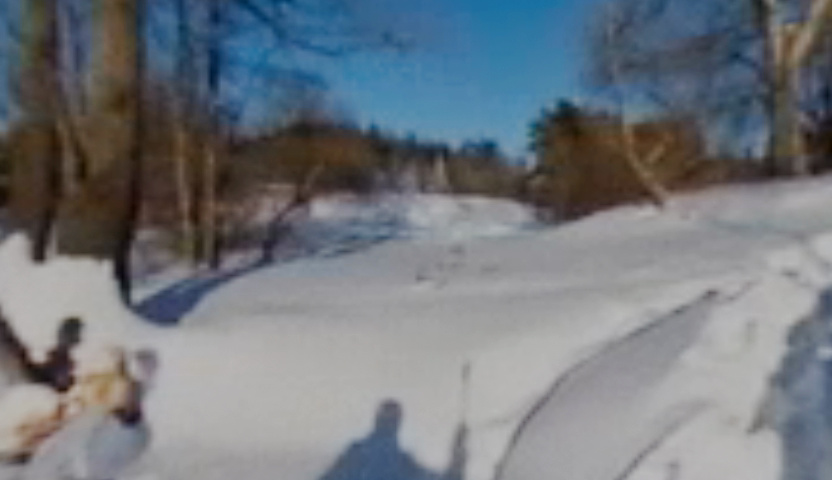} &
\includegraphics[height=\lh]{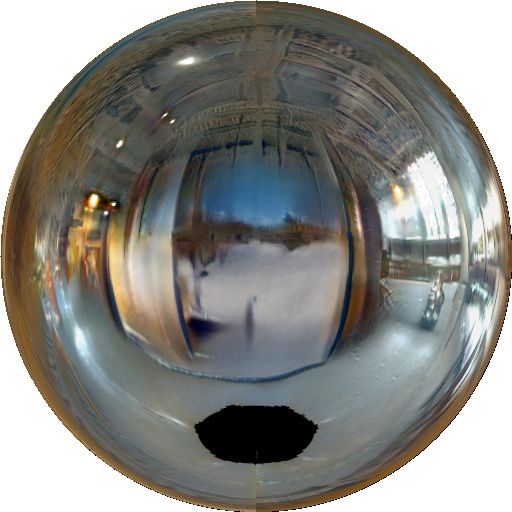} &
\includegraphics[height=\lh]{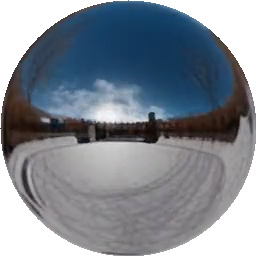} &
\includegraphics[height=\lh]{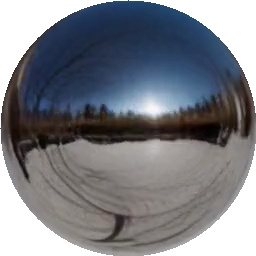} &
\includegraphics[height=\lh]{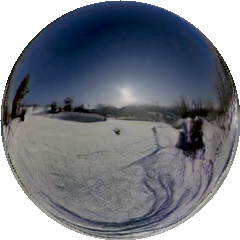} &
\includegraphics[height=\lh]{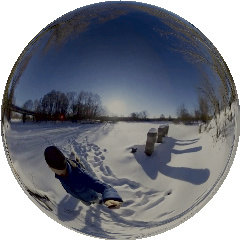} \\

\hdashline

\includegraphics[height=\lh]{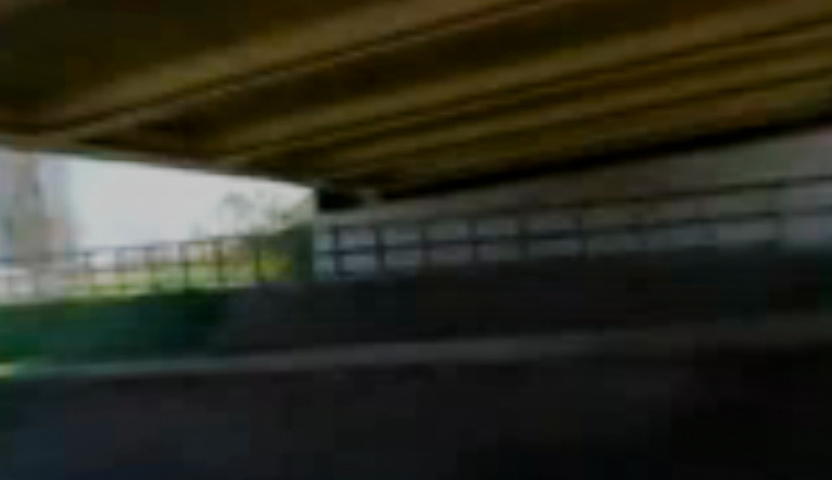} &
\includegraphics[height=\lh]{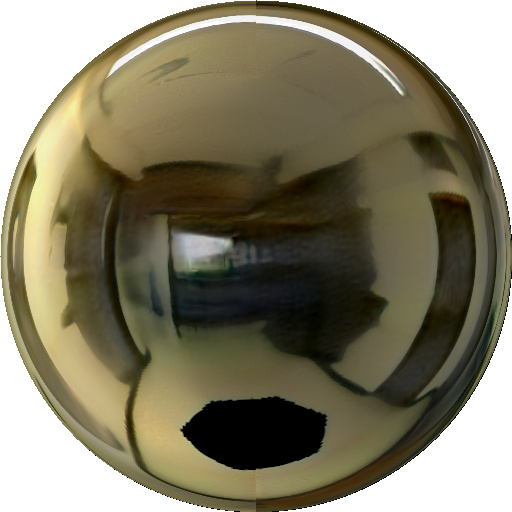} &
\includegraphics[height=\lh]{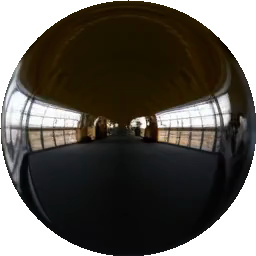} &
\includegraphics[height=\lh]{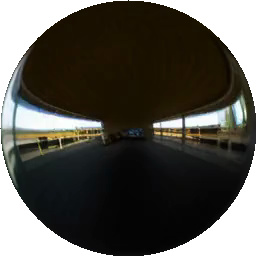} &
\includegraphics[height=\lh]{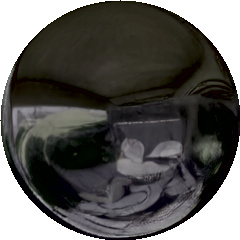} &
\includegraphics[height=\lh]{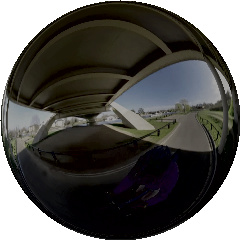} &
\rinput{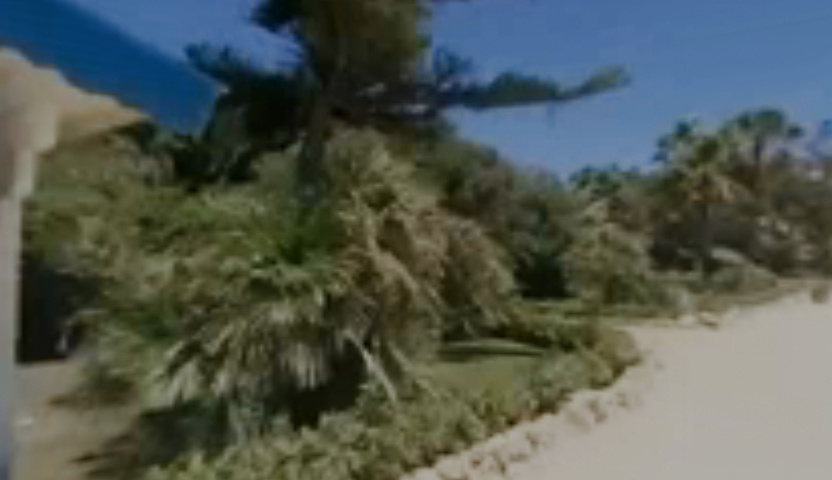} &
\includegraphics[height=\lh]{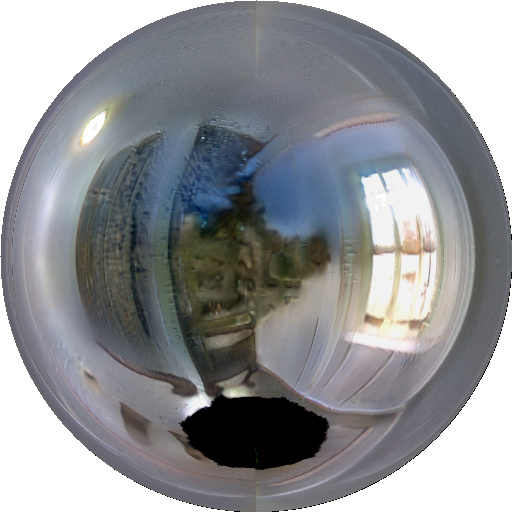} &
\includegraphics[height=\lh]{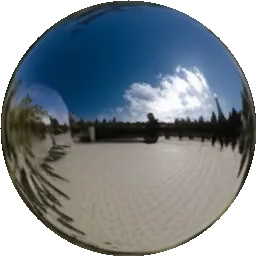} &
\includegraphics[height=\lh]{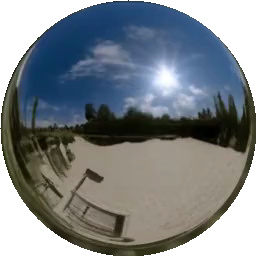} &
\includegraphics[height=\lh]{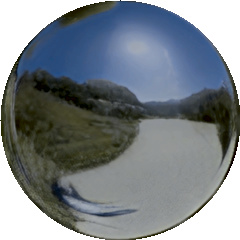} &
\includegraphics[height=\lh]{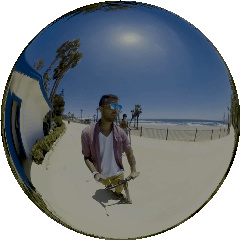} \\
\includegraphics[height=\lh]{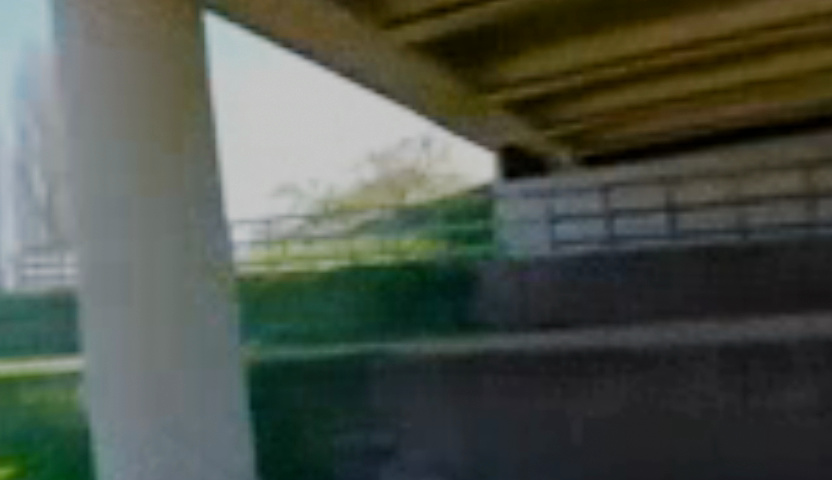} &
\includegraphics[height=\lh]{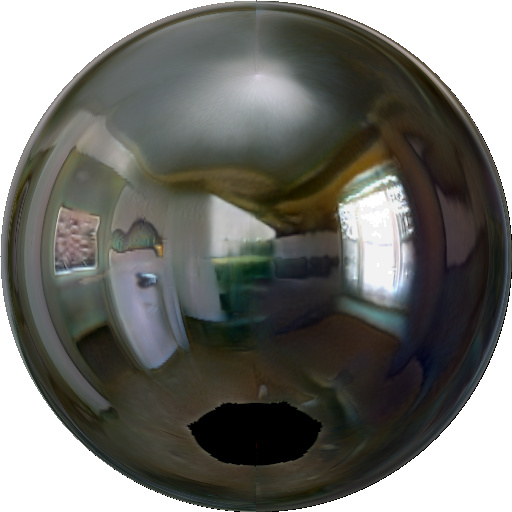} &
\includegraphics[height=\lh]{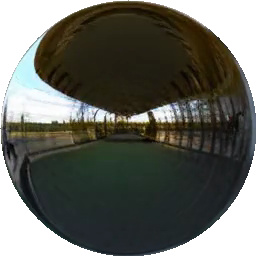} &
\includegraphics[height=\lh]{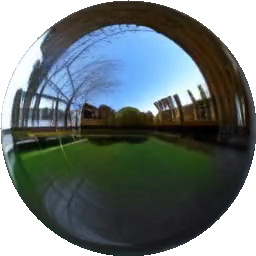} &
\includegraphics[height=\lh]{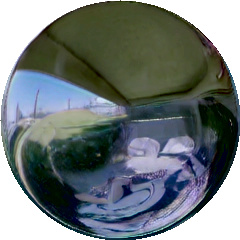} &
\includegraphics[height=\lh]{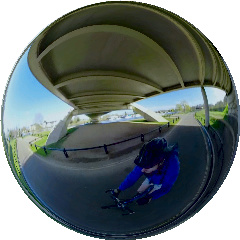} &
\rinput{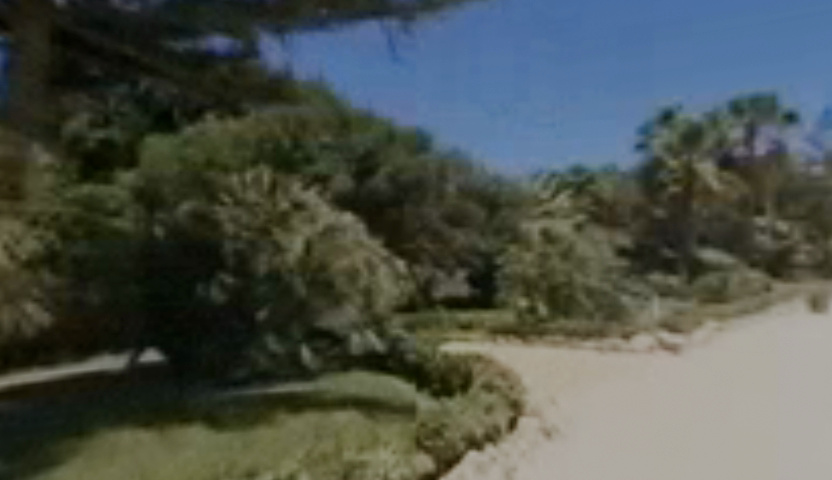} &
\includegraphics[height=\lh]{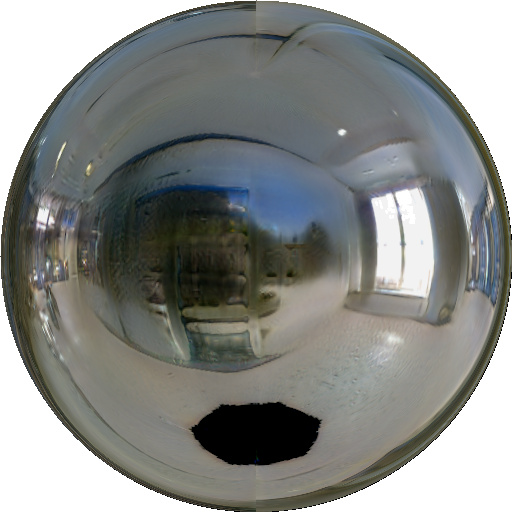} &
\includegraphics[height=\lh]{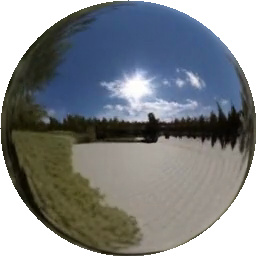} &
\includegraphics[height=\lh]{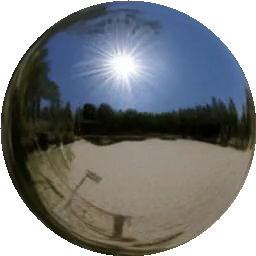} &
\includegraphics[height=\lh]{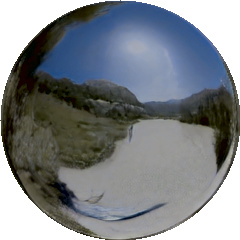} &
\includegraphics[height=\lh]{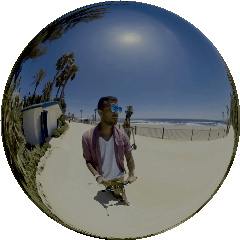} \\
\includegraphics[height=\lh]{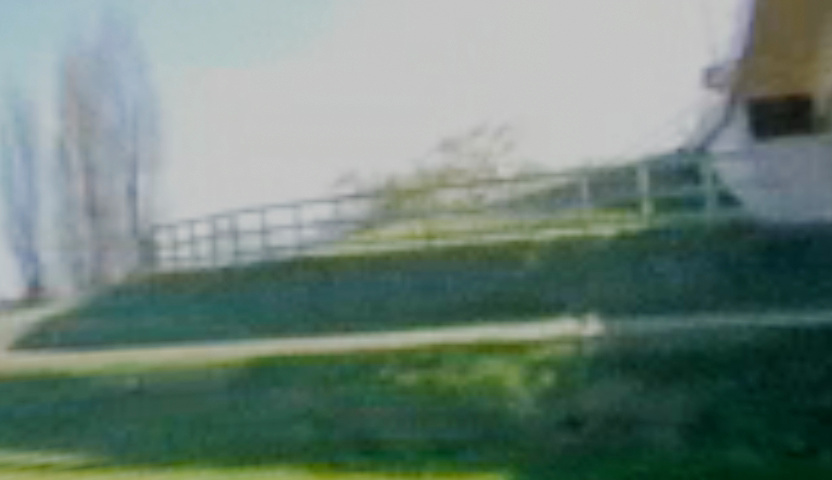} &
\includegraphics[height=\lh]{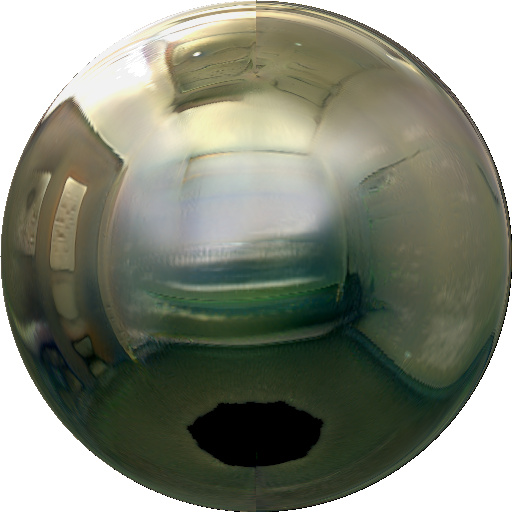} &
\includegraphics[height=\lh]{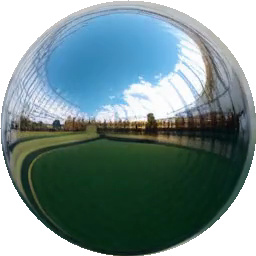} &
\includegraphics[height=\lh]{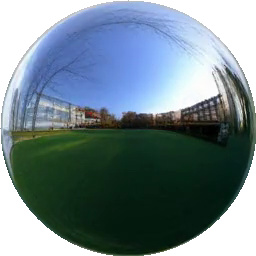} &
\includegraphics[height=\lh]{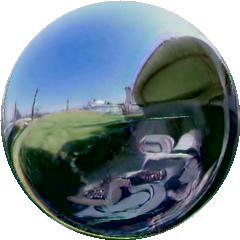} &
\includegraphics[height=\lh]{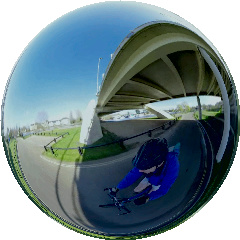} &
\rinput{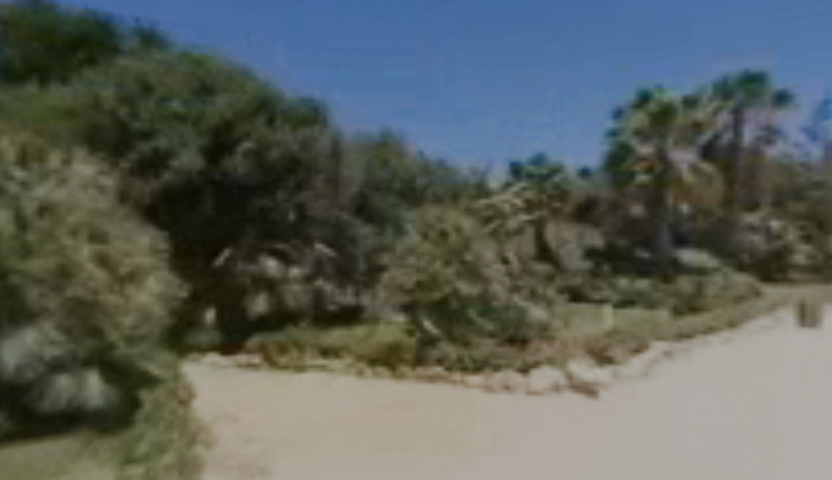} &
\includegraphics[height=\lh]{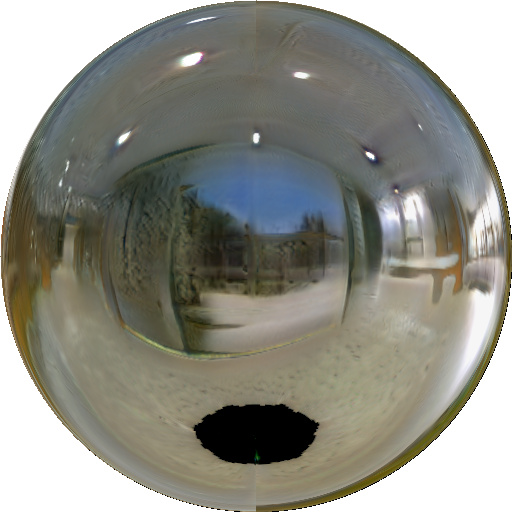} &
\includegraphics[height=\lh]{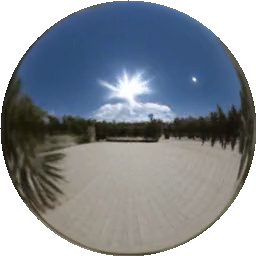} &
\includegraphics[height=\lh]{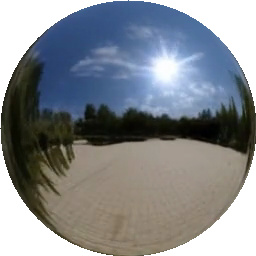} &
\includegraphics[height=\lh]{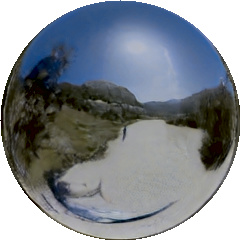} &
\includegraphics[height=\lh]{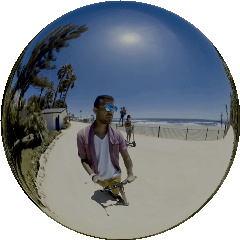} \\
\includegraphics[height=\lh]{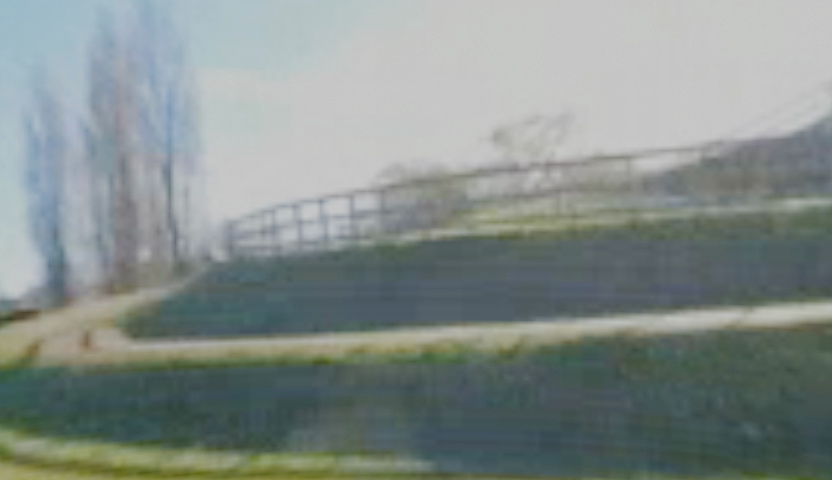} &
\includegraphics[height=\lh]{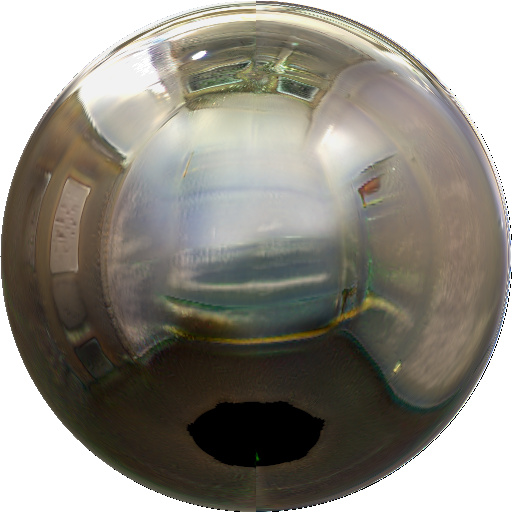} &
\includegraphics[height=\lh]{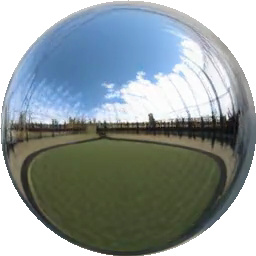} &
\includegraphics[height=\lh]{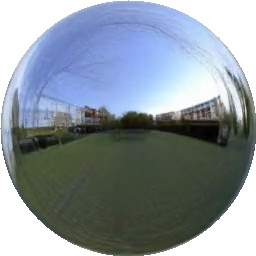} &
\includegraphics[height=\lh]{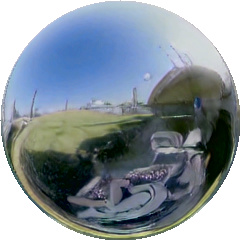} &
\includegraphics[height=\lh]{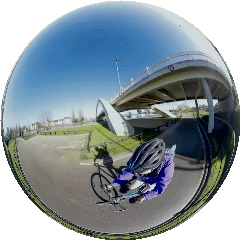} &
\rinput{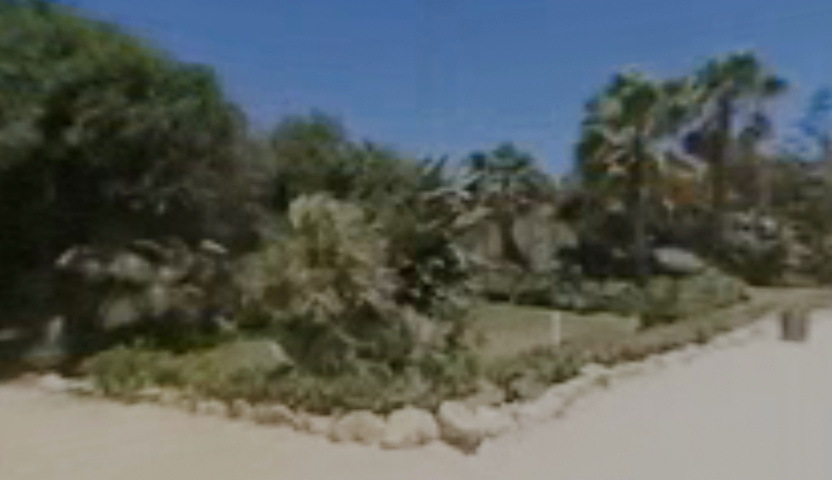} &
\includegraphics[height=\lh]{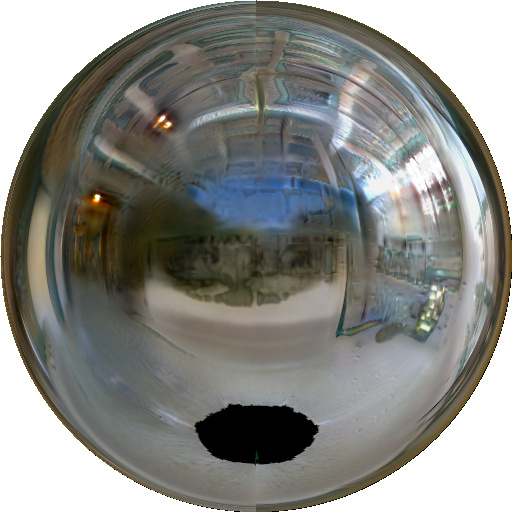} &
\includegraphics[height=\lh]{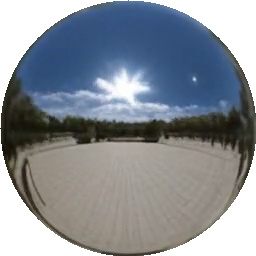} &
\includegraphics[height=\lh]{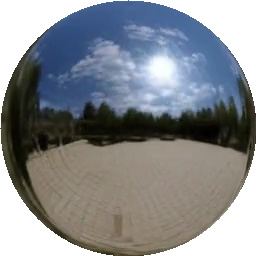} &
\includegraphics[height=\lh]{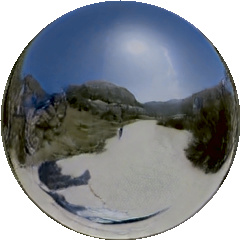} &
\includegraphics[height=\lh]{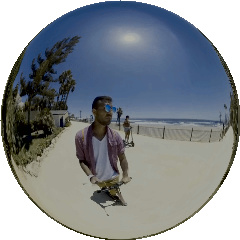} \\[3pt]
\multicolumn{1}{c}{{\scriptsize Input}} &
\multicolumn{1}{c}{{\scriptsize SL~\cite{wang2022stylelight}}} &
\multicolumn{1}{c}{{\scriptsize DL~\cite{phongthawee2024diffusionlight}}} &
\multicolumn{1}{c}{{\scriptsize DLT~\cite{chinchuthakun2025diffusionlightturbo}}} &
\multicolumn{1}{c}{{\scriptsize Ours}} &
\multicolumn{1}{c}{{\scriptsize GT}} &
\multicolumn{1}{c}{{\scriptsize Input}} &
\multicolumn{1}{c}{{\scriptsize SL~\cite{wang2022stylelight}}} &
\multicolumn{1}{c}{{\scriptsize DL~\cite{phongthawee2024diffusionlight}}} &
\multicolumn{1}{c}{{\scriptsize DLT~\cite{chinchuthakun2025diffusionlightturbo}}} &
\multicolumn{1}{c}{{\scriptsize Ours}} &
\multicolumn{1}{c}{{\scriptsize GT}} \\
\end{tabular}
\caption{Qualitative comparison with baseline lighting estimation models. We compare our method against three methods: StyleLight~\cite{wang2022stylelight} (SL), DiffusionLight~\cite{phongthawee2024diffusionlight} (DL), and DiffusionLightTurbo~\cite{chinchuthakun2025diffusionlightturbo} (DLT). For each block, the first row shows input video frames, followed by the predicted lighting conditions from each method, and ground truth (GT) in the final row. Our method produces lighting estimates that closely match the ground truth in terms of intensity and direction, while maintaining temporal consistency.
}
\label{fig:qualitative_comparison}
\end{figure*}

\subsection{Qualitative Evaluation}

We provide a qualitative comparison with state-of-the-art image relighting methods in \cref{fig:qualitative_comparison}. The baselines including DiffusionLight~\cite{phongthawee2024diffusionlight}, DiffusionLight-Turbo~\cite{chinchuthakun2025diffusionlightturbo}, and StyleLight~\cite{wang2022stylelight}, are designed for single-image manipulation. To adapt them for video evaluation, we apply each method independently to each frame of source video sequences. As illustrated, this per-frame approach leads to significant temporal artifacts. The baseline methods exhibit noticeable flickering, where the intensity and color of highlights and shadows change erratically between consecutive frames. This is an expected outcome, as these models lack any inherent mechanism to enforce temporal consistency.

\begin{figure*}[t]
    \centering
    \includegraphics[width=\linewidth]{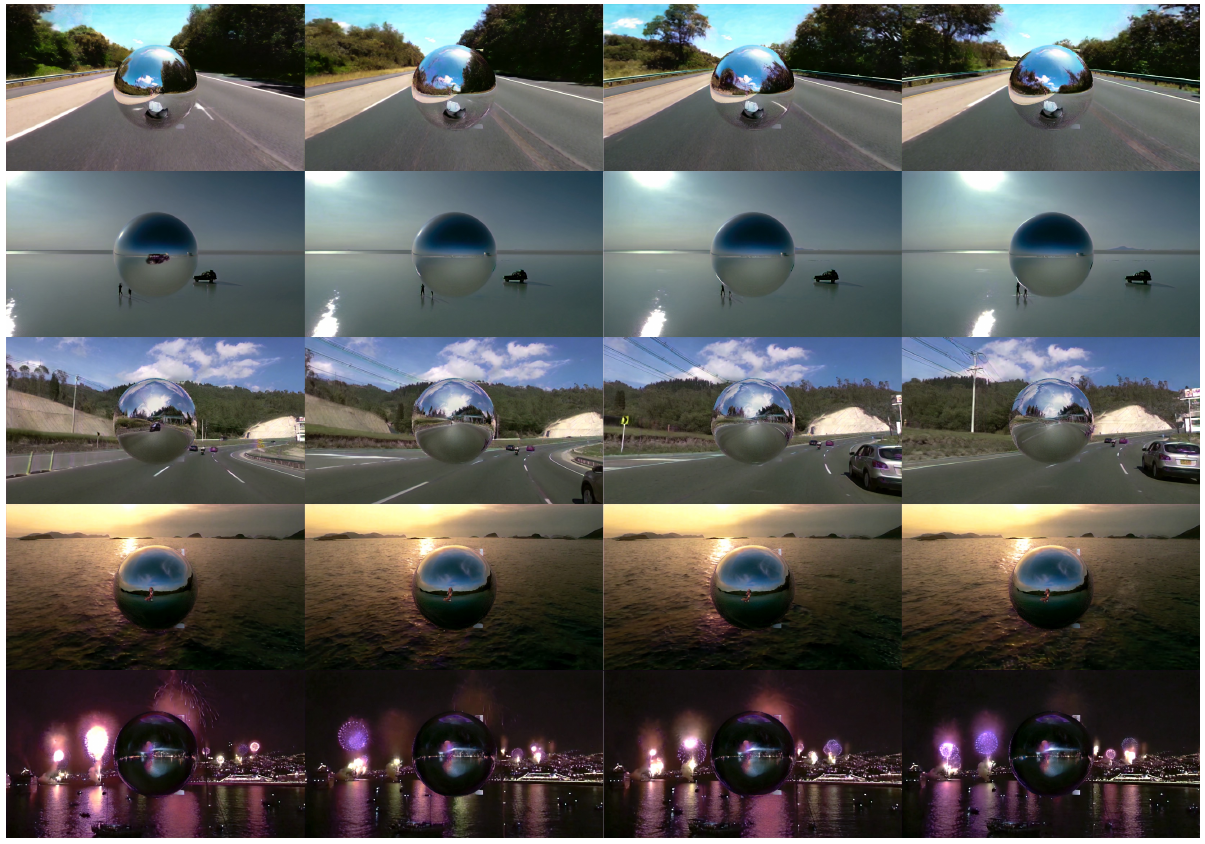}
    \caption{\rev{Qualitative results on in-the-wild videos. For visualization purposes, all frames are tone-mapped from their original HDR format.}}
    \label{fig:wild}
\end{figure*}

\rev{\noindent\textbf{Generalization to in-the-wild videos.}
As shown in \cref{fig:wild}, V-LITE generalizes to diverse in-the-wild videos under both day and night conditions, producing spatially accurate and temporally coherent lighting estimates even in complex real-world footage.}

\subsection{Ablation Study}

To validate our key design choices, we conduct a set of ablation experiments that isolate the contributions of the HDR-aware VAE, the LoRA-based fine-tuning procedure, and our mixed-data training strategy. We evaluate three ablated variants against our full model and report the quantitative results in \cref{tab:main_quantitative_results}.

\noindent\textbf{LDR baseline.} We first evaluate the original pretrained model of Jiang \etal~\cite{jiang2025vace} without both our HDR VAE and the fine-tuning stage. Although this LDR baseline can inpaint chrome balls in videos, it cannot generate HDR content. To enable a comparison with HDR-capable models, we apply the inverse tone-mapping derived from the HDR ground truth to lift its LDR predictions into the HDR domain. The results support our claim that while video diffusion models inherently encode lighting information, they lack the representational precision to express it in the HDR space.

\noindent\textbf{Frozen backbone.} We incorporate our pretrained HDR VAE directly into the baseline model without any task-specific fine-tuning. This configuration introduces a domain mismatch between the original LDR training distribution and the new HDR latent space, significantly degrading the generative performance.

\noindent\textbf{Video-only.} To evaluate the impact of our mixed-data training strategy, we fine-tune the model equipped with the HDR VAE exclusively on the video dataset. While this variant aligns the generative priors to the video inpainting task, relying solely on tonemapped HDR video data limits the model's ability to predict accurate physical HDR values.

Instead, our full model incorporates both the HDR VAE and the LoRA-based fine-tuning, utilizing a mixed-data training strategy. It outperforms all ablation variants, demonstrating that all proposed components are indispensable.

\begin{figure*}[tb]
\def\lw{0.2\linewidth}
\def\hlw{0.08333333\linewidth}
\def\ftsz{\normalsize}
\renewcommand\tabcolsep{0.0pt}
\renewcommand{\arraystretch}{0}
\centering \small
\begin{tabular}{cccccc}
\includegraphics[width=\lw]{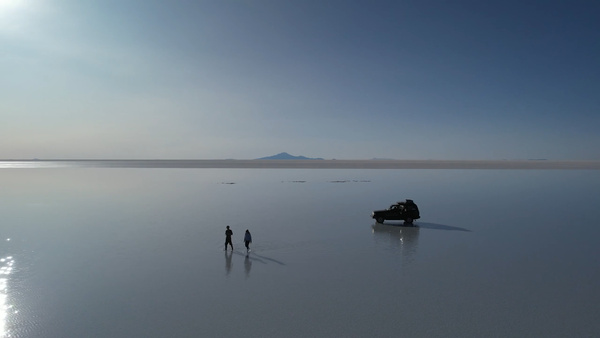} &
\includegraphics[width=\lw]{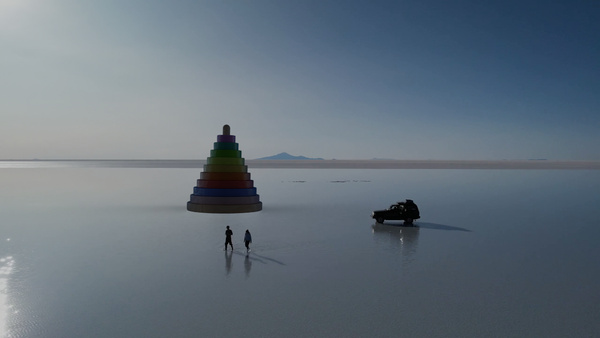} &
\includegraphics[width=\lw]{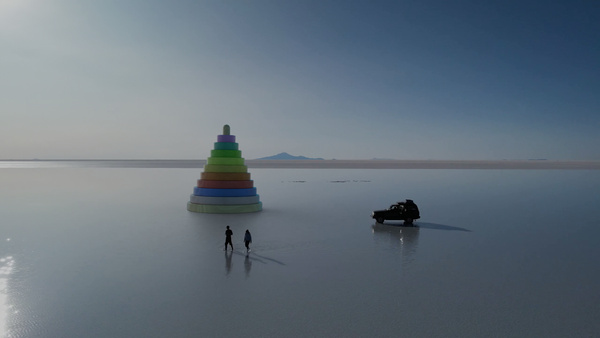} &
\includegraphics[width=\lw]{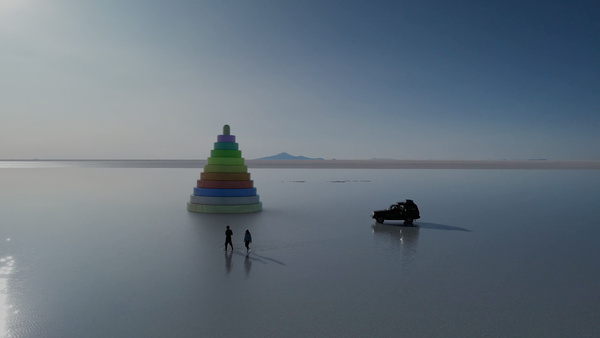} &
\includegraphics[width=\lw]{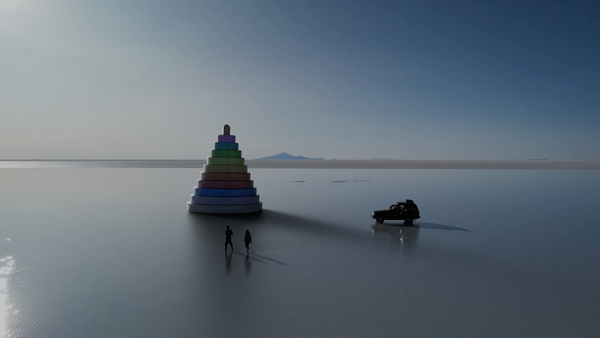} \\
\includegraphics[width=\lw]{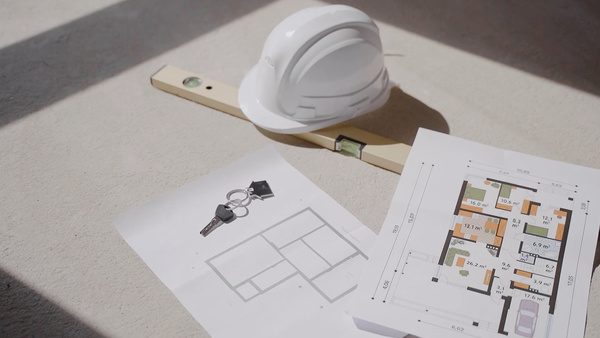} &
\includegraphics[width=\lw]{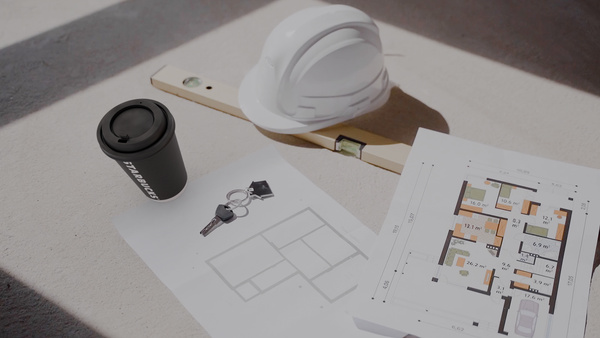} &
\includegraphics[width=\lw]{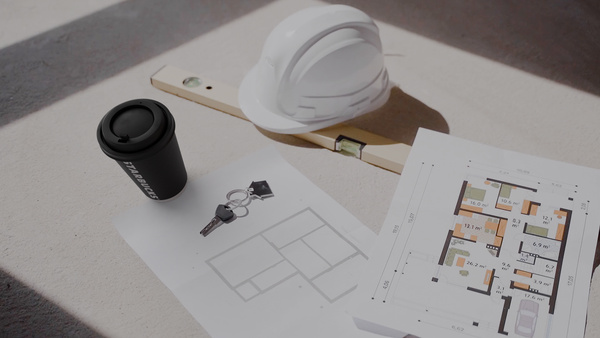} &
\includegraphics[width=\lw]{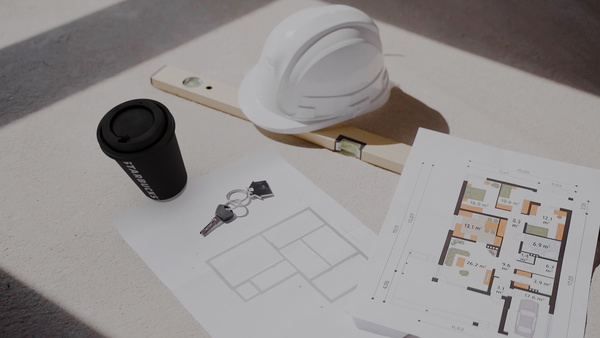} &
\includegraphics[width=\lw]{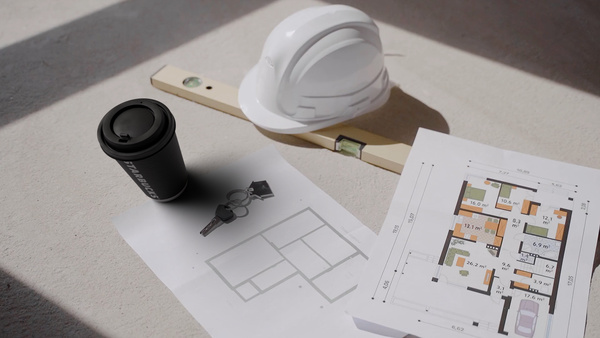} \\[3pt]
Input & StyleLight~\cite{wang2022stylelight} & DL~\cite{phongthawee2024diffusionlight}&DLT~\cite{chinchuthakun2025diffusionlightturbo}& Ours
\end{tabular}
\caption{Qualitative comparison of virtual object insertion. We compare our method against  StyleLight~\cite{wang2022stylelight}, DiffusionLight~\cite{phongthawee2024diffusionlight}, and DiffusionLightTurbo~\cite{chinchuthakun2025diffusionlightturbo}. To ensure a fair comparison, we use the dynamic HDR environment map sequence estimated by each method to illuminate a virtual object within a fixed 3D scene. The first column displays the input video frame, followed by the rendered results from each competing method and our own. Our estimated lighting produces a more physically plausible result, enabling seamless integration of the virtual object. Note the superior quality of the specular highlights and the accurate rendering of both hard and soft cast shadows.}
\label{fig:insertion}
\end{figure*}

\subsection{Virtual Object Insertion}

We demonstrate the effectiveness of our estimated HDR environment maps by inserting virtual objects into real-world videos. We first process the input LDR video to solve for the 3D camera motion and establish a ground plane using built-in motion tracker from Blender~\cite{blender}. The dynamic HDR environment map sequence generated by V-LITE is then used as the exclusive lighting source for a virtual object. This object is rendered from the solved camera's perspective and composited back onto the original video. The results in \cref{fig:insertion} show that the virtual objects are seamlessly integrated, with lighting and shadows that are both physically plausible and temporally consistent.

To rigorously validate perceptual realism and lighting consistency, we conducted a user study with 10 participants across 20 diverse videos. For perceptual realism, 64\% of responses rated our insertions as ``Perfect'', and 28\% as ``Acceptable''. For lighting consistency, 66.5\% rated the results as ``Perfect'', with less than 2\% of total responses considered failures.

\subsection{Discussion}

\noindent\textbf{Tonemapped HDR videos and mixed-data training.}
Capturing in-the-wild HDR videos requires specialized multi-exposure cameras, which fundamentally limits dataset scale and scene diversity. To learn complex dynamic lighting, our V-LITESet utilizes computationally tonemapped HDR videos. While these are not photometrically perfect HDRs, our rigorous VLM-based filtering explicitly discards 57\% of heavily clipped data to ensure reasonable intensity ratios. By mixing these dynamic tonemapped videos with static real-world HDR images, V-LITE effectively integrates absolute photometric priors to overcome the physical scarcity of real-world HDR video data.

\noindent\textbf{Ill-posed estimation and visual trade-offs.}
Estimating an HDR environment from a single LDR observation is an inherently ill-posed problem, as the original tone-mapping process is non-invertible. Because V-LITE leverages the generative priors of diffusion models to naturally resolve this ambiguity rather than performing a strict pixel-to-pixel translation, the output may exhibit slight background visual shifts (\eg, lower contrast on certain bright surfaces). These are expected physical trade-offs for accurately extracting HDR reflections and dominant light directions. As evidenced by our high-fidelity virtual object insertions, V-LITE robustly estimates practical dynamic lighting.

\begin{figure}[tb]
    \centering
    \includegraphics[width=\columnwidth]{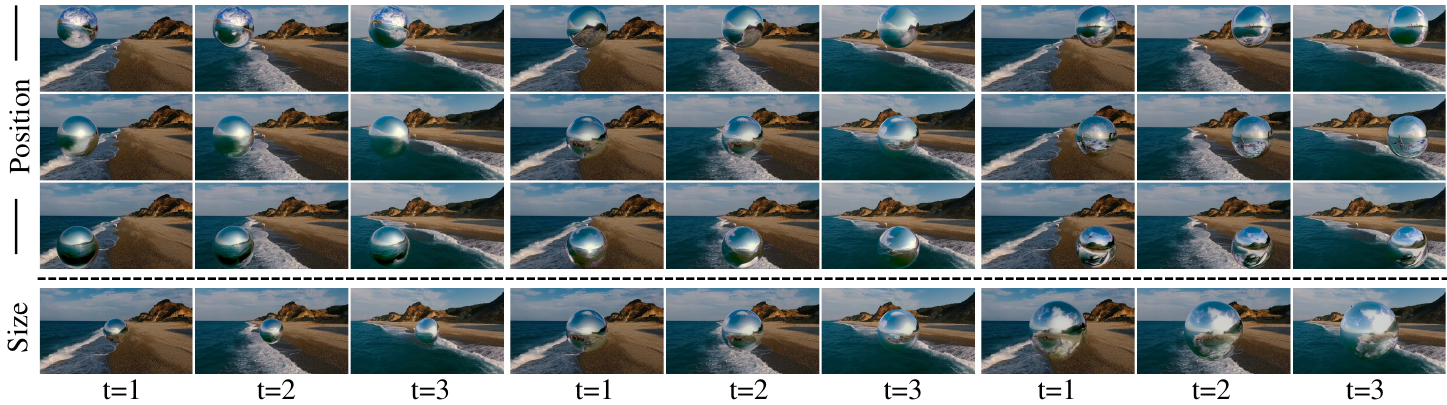}
    \caption{Visualization of zero-shot generalization by varying probe position and size.}
    \label{fig:spatially}
\end{figure}

\noindent\textbf{Zero-shot generalization and spatial variance.}
We further examine the influence of the probe's position and size, observing that the inpainted reflections dynamically respond to its spatial positioning. As depicted in \cref{fig:spatially}, rows 1--3 illustrate off-center placements from top-left to top-right, while row 4 explores different mask coverage ratios (30\%, 50\%, and 70\%). All preserve structural integrity and realistic illumination, implying that V-LITE inherently captures underlying scene geometry.

\section{Conclusion}
We present V-LITE, a unified HDR-aware framework that produces dynamic HDR environment maps from in-the-wild videos. We also introduce V-LITESet, a hybrid dataset of tonemapped HDR videos and real-world HDR images curated to advance video-based lighting estimation. Our formulation natively treats lighting recovery as a latent video-inpainting problem, letting the model fully exploit the generative priors of video diffusion models. Experiments show that V-LITE delivers high-fidelity, temporally coherent environment maps in challenging dynamic scenes.

\begin{figure}[tb]
    \centering
    \includegraphics[width=1\linewidth]{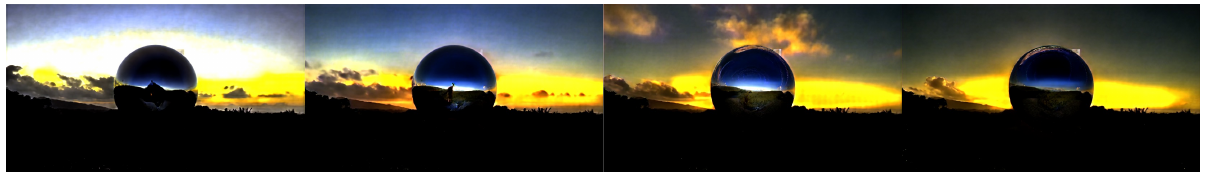}
    \caption{Visualization of a failure case. When encountering extreme, out-of-distribution inputs, the model may occasionally produce mismatched environment maps.}
    \label{fig:failure}
\end{figure}

\noindent\textbf{Limitations and failure cases.}
While V-LITE demonstrates strong generalization, its data-driven nature makes it sensitive to out-of-distribution inputs. As illustrated in \cref{fig:failure}, videos with extreme illumination conditions can occasionally lead to mismatched environment maps. Additionally, since our underlying video backbones~\cite{wan2025wan, yang2024cogvideox} struggle with long-duration generation, V-LITE's ability to estimate temporally stable lighting for extended sequences remains constrained.

\section*{Acknowledgment}
This work is supported by National Natural Science Foundation of China (Grant No. 62136001) and Beijing Major Science and Tech\-nology Project (Grant No. Z251100008\allowbreak125009). PKU-affiliated authors thank openbayes.com for providing computing resources.

\bibliographystyle{splncs04}
\bibliography{main}
\end{document}